\pdfoutput=1
\documentclass[11pt,a4paper]{article}
\usepackage[dvipsnames]{xcolor}
\usepackage[hyperref]{acl2021}
\usepackage{times}
\usepackage{latexsym}
\usepackage{inconsolata}
\usepackage{microtype}
\usepackage[fleqn]{amsmath}
\usepackage{amsfonts}
\usepackage{url}
\usepackage{booktabs}
\usepackage{multirow}
\usepackage{xspace}
\usepackage{bm}
\usepackage{tikz}
\usepackage{tikz-dependency}
\usepackage{tabularx}
\usepackage[T5,T1]{fontenc}
\usepackage[utf8]{inputenc}
\usepackage[title]{appendix}
\usepackage{bibunits}
\usepackage{fancybox}
\usepackage{amsthm}
\usepackage{mathtools}
\usepackage{amssymb}

\usetikzlibrary{shapes.misc, fit, positioning, calc}

\tikzset{
    vertex/.style={circle,draw,minimum size=1.5em},
    edge/.style={->,> = latex'}
}

\aclfinalcopy %
\def\aclpaperid{***} %

\title{
  Transition-based Bubble Parsing:\\
  Improvements on Coordination Structure Prediction
}

\author{Tianze Shi\\
  Cornell University\\
  \texttt{tianze@cs.cornell.edu}\\\And
  Lillian Lee\\
  Cornell University\\
  \texttt{llee@cs.cornell.edu} \\
}

\date{}

\newtoggle{comment}
\toggletrue{comment}
\mathchardef\mhyphen="2D

\newcommand{\cond}[1]{\MakeUppercase #1:\xspace} %
\newcommand{\rootsym}{\ensuremath{\texttt{RT}}\xspace}

\def\Snospace~{\S{}}

\newtheorem*{theorem*}{Theorem}
\newtheorem{lemma}{Lemma}
\newtheorem*{lemma*}{Lemma}

\newenvironment{sproof}{
  \proof
}{\endproof}

\newcommand{\transition}[1]{\textsc{#1}}
\newcommand{\shift}{\transition{Shift}}
\newcommand{\leftarc}{\transition{LeftArc}}
\newcommand{\rightarc}{\transition{RightArc}}
\newcommand{\bubbleopen}{\transition{BubbleOpen}}
\newcommand{\bubbleattach}{\transition{BubbleAttach}}
\newcommand{\bubbleclose}{\transition{BubbleClose}}

\newcommand{\stack}{\ensuremath{\sigma}}
\newcommand{\buffer}{\ensuremath{\beta}}
\newcommand{\stackelement}[1]{\ensuremath{s_{#1}}}
\newcommand{\bufferelement}[1]{\ensuremath{b_{#1}}}

\newcommand{\bubbleset}{\ensuremath{\mathcal{B}}}

\newcommand{\openset}{\ensuremath{\mathcal{O}}\xspace}
\newcommand{\emptybuffer}{\ensuremath{\varnothing}}

\newcommand{\archybrid}{Arc-Hybrid\xspace}
\newcommand{\transitionsystem}{Bubble-Hybrid\xspace}

\newcommand{\taketransition}[1]{\ensuremath{\xRightarrow{#1\,\,}}}

\newcommand{\partialmapsto}{\ensuremath{\mapstochar\rightharpoonup}}

\newcommand{\extendfunction}{\Cup}

\newcolumntype{R}[1]{>{\RaggedLeft\arraybackslash}m{#1}}
\newcolumntype{S}{>{\small}l}

\newcommand{\posscite}[1]{\citeauthor{#1}'s \citeyearpar{#1}}

\newcommand{\bivec}[1]{\ensuremath{\mathbf{#1}}}

\newcommand{\wvec}[1]{\ensuremath{\bivec{w}_{#1}}}
\newcommand{\vvec}[1]{\ensuremath{\bivec{v}_{#1}}}

\newcommand{\resnumber}[1]{\ensuremath{#1}}

\defcitealias{hara+09}{HSOM09}
\defcitealias{ficler-goldberg16}{FG16}
\defcitealias{teranishi+17}{TSM17}
\defcitealias{teranishi+19}{TSM19}

\newcommand{\bert}{\texttt{BERT}\xspace}
\newcommand{\bertbase}{\texttt{BERT}\textsubscript{base}\xspace}

\begin{document}
\maketitle
\begin{abstract}
We propose a transition-based bubble parser
to perform coordination structure identification
and dependency-based syntactic analysis simultaneously.
Bubble representations were proposed in
the formal linguistics literature decades ago;
they enhance dependency trees by
encoding coordination boundaries and internal
relationships within coordination structures
explicitly.
In this paper,
we introduce a transition system and neural models
for parsing these bubble-enhanced structures.
Experimental results on the English Penn Treebank and
the English GENIA corpus
show that our parsers beat previous state-of-the-art approaches
on the task of coordination structure prediction,
especially for the subset of sentences with complex coordination structures.\footnote{
Code at \url{github.com/tzshi/bubble-parser-acl21}.
}
\end{abstract}

\begin{bibunit}[acl_natbib]

\section{Introduction}
\label{sec:intro}

Coordination structures are prevalent in treebank data \citep{ficler-goldberg16b},
especially in long sentences \citep{kurohashi-nagao94a},
and they are among the most challenging constructions for NLP models.
Difficulties in correctly identifying coordination structures
have consistently contributed to a significant portion of errors
in state-of-the-art parsers \citep{collins03,goldberg-elhadad10,ficler-goldberg17}.
These errors can
further propagate to downstream NLP modules and applications,
and limit their performance and utility.
For example, \citet{saha+17} report that
missing conjuncts account for two-thirds of
the errors in recall
made by their open information extraction system.

Coordination constructions are particularly challenging
for the widely-adopted dependency-based
paradigm of
syntactic analysis,
since the asymmetric definition of
head-modifier dependency relations
is not directly compatible with the symmetric nature of
the relations among the participating conjuncts and coordinators.\footnote{
   \citet{rambow-2010-simple} comments on other divergences between syntactic representation
   and  syntactic phenomena.
}
Existing treebanks usually resort to introducing special relations
to represent coordination structures.
But, there remain theoretical and empirical challenges
regarding how to most effectively encode
information like modifier sharing relations
while still permitting accurate statistical syntactic analysis.

\begin{figure}[t]
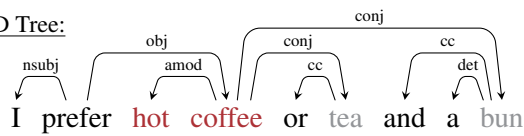


{\small \underline{Bubble Tree:}}
\vspace{-10pt}
\begin{center}

\begin{dependency}[theme=simple, label style={font=\footnotesize}]%
\begin{deptext}[column sep=0.1cm]
I \& prefer \& \textcolor{Maroon}{hot} \& \textcolor{Maroon}{coffee} \& or \& \textcolor{Maroon}{tea} \& and \& a \& \textcolor{CadetBlue}{bun} \\
\end{deptext}

\wordgroup[style={draw=Maroon, rounded corners=6pt, inner sep=1pt, anchor=south, minimum height=30pt}, yshift=-11pt]{1}{4}{6}{ccp1}

\wordgroup[style={draw=white}]{1}{4}{4}{coord11}
\node (coord11n) at ($ (coord11) + (-0.15, 0.89) $) {};
\node (coord11s) at ($ (coord11) + (-0.15, 0.1) $) {};
\draw[edge] (coord11n) -- (coord11s) node (coord11lbl) [midway, anchor=west, xshift=-3pt] {\scriptsize{conj}};

\wordgroup[style={draw=white}]{1}{5}{5}{cc1}
\node (cc1n) at ($ (cc1) + (-0.05, 0.89) $) {};
\node (cc1s) at ($ (cc1) + (-0.05, 0.1) $) {};
\draw[edge] (cc1n) -- (cc1s) node (cc1lbl) [midway, anchor=west, xshift=-3pt] {\scriptsize{cc}};

\wordgroup[style={draw=white}]{1}{6}{6}{coord12}
\node (coord12n) at ($ (coord12) + (-0.15, 0.89) $) {};
\node (coord12s) at ($ (coord12) + (-0.15, 0.1) $) {};
\draw[edge] (coord12n) -- (coord12s) node (coord12lbl) [midway, anchor=west, xshift=-3pt] {\scriptsize{conj}};

\wordgroup[style={draw=white}]{1}{3}{3}{hot}

\groupedge[edge height=50pt, edge start x offset=-15pt, edge vertical padding=-2.5pt]{ccp1}{hot}{\hspace{-10pt}amod}{20pt}

\wordgroup[style={rounded corners=6pt, inner sep=1pt, anchor=south, minimum height=44pt}, yshift=-13pt]{1}{3}{9}{ccp2}
\node (coord21n) at ($ (ccp1) + (-0.15, 1.08) $) {};
\node (coord21s) at ($ (ccp1) + (-0.15, 0.4) $) {};
\draw[edge] (coord21n) -- (coord21s) node (coord21lbl) [midway, anchor=west, xshift=-3pt] {\scriptsize{conj}};

\wordgroup[style={draw=white}]{1}{7}{7}{cc2}
\node (cc2n) at ($ (cc2) + (-0.05, 1.31) $) {};
\node (cc2s) at ($ (cc2) + (-0.05, 0.1) $) {};
\draw[edge] (cc2n) -- (cc2s) node (cc2lbl) [midway, anchor=west, xshift=-3pt] {\scriptsize{cc}};

\wordgroup[style={draw=white}]{1}{9}{9}{coord22}
\node (coord22n) at ($ (coord22) + (-0.15, 1.3) $) {};
\node (coord22s) at ($ (coord22) + (-0.15, 0.1) $) {};
\draw[edge] (coord22n) -- (coord22s) node (cc2lbl) [midway, anchor=west, xshift=-3pt] {\scriptsize{conj}};

\depedge[segmented edge, edge unit distance=1.8ex, edge start x offset=-3pt]{9}{8}{det}
\depedge[segmented edge, edge unit distance=1.8ex]{2}{1}{nsubj}

\wordgroup[style={draw=white}]{1}{2}{2}{prefer}

\groupedge[edge start x offset=-5pt, edge vertical padding=-2.5pt, edge end x offset=-60pt]{prefer}{ccp2}{\hspace{-15pt}obj}{0pt}

\end{dependency}
\end{center}

{\small \underline{UD Tree:}}
\vspace{-18pt}
\begin{center}

\begin{dependency}[theme=simple, label style={font=\footnotesize}, segmented edge, edge unit distance=1.8ex]
\begin{deptext}[column sep=0.1cm]
I \& prefer \& \textcolor{Maroon}{hot} \& \textcolor{Maroon}{coffee} \& or \& \textcolor{Gray}{tea} \& and \& a \& \textcolor{Gray}{bun} \\
\end{deptext}
\depedge{2}{1}{nsubj}
\depedge{2}{4}{obj}
\depedge{4}{3}{amod}
\depedge[edge start x offset=5pt]{4}{6}{conj}
\depedge{6}{5}{cc}
\depedge[edge unit distance=1.1ex]{4}{9}{conj}
\depedge{9}{7}{cc}
\depedge[edge start x offset=-3pt]{9}{8}{det}
\end{dependency}
\end{center}

\caption{
Bubble tree 
and (basic) UD tree for the same example sentence.
(For clarity, we omit  punctuation and single-word bubble boundaries.)
Bubbles  
explicitly encode
the scope of the shared modifier \emph{hot} with respect to the  nested coordination,
whereas the UD tree 
gives both \emph{tea} and \emph{bun} identical relationships to \emph{hot}.
}
\label{fig:main-ex}

\end{figure}

In this paper, we explore \posscite{kahane97} alternative solution:
extend the dependency-tree representation
by introducing {\em bubble structures}
to explicitly encode coordination boundaries.
The co-heads within a bubble
enjoy a symmetric relationship,
as befits a model of conjunction.
Further, bubble trees support representation of nested coordination,
with the scope of shared modifiers identifiable
by the attachment sites of bubble arcs.
\autoref{fig:main-ex} compares a bubble tree
against a Universal Dependencies \cite[UD;][]{nivre+16,nivre+20} tree
for the same sentence.

Yet, despite theses advantages,
implementation of the formalism was not broadly pursued, for reasons unknown to us.
Given its appealing and intuitive treatment of coordination phenomena,
we revisit the bubble tree formalism, introducing and implementing
a transition-based solution for parsing bubble trees.
Our transition system, \transitionsystem, extends the \archybrid transition system \citep{kuhlmann+11a}
with three bubble-specific transitions,
each corresponding to opening, expanding, and closing bubbles.
We show that our transition system is both sound and complete
with respect to projective bubble trees (defined in \autoref{sec:bubble}).

Experiments on the English Penn Treebank \citep[PTB;][]{marcus+93}
extended with coordination annotation \citep{ficler-goldberg16b}
and the English GENIA treebank \citep{kim+03}
demonstrate the effectiveness of our proposed transition-based bubble parsing
on the task of coordination structure prediction.
Our method achieves state-of-the-art performance on both datasets
and improves accuracy
on the subset of sentences exhibiting complex coordination structures.

\section{Dependency Trees and Bubble Trees}
\label{sec:repr}

\subsection{Dependency-based Representations for Coordination Structures}
\label{sec:deptree}

A dependency tree encodes syntactic relations
via directed bilexical dependency edges.
These are natural for representing argument and adjunct modification,
but \citet{popel+13} point out that
``dependency representation is at a loss when it comes to representing
paratactic linguistic phenomena such as coordination, whose nature is
symmetric (two or more conjuncts play the same role), as opposed to the
head-modifier asymmetry of dependencies'' (pg. 517).

If one nonetheless persists in using dependency relations to annotate all syntactic structures,
as is common practice in most dependency treebanks \citep[][\emph{inter alia}]{hajic+01,nivre+16},
then one must
introduce special relations to represent coordination structures
and promote one element from each coordinated phrase
to become the ``representational head''.
One choice is to specify one of the conjuncts
as the ``head'' \cite{melcuk88,melcuk03,jarvinen-tapanainen98,lombardo-lesmo98}
(e.g., in \autoref{fig:main-ex}, the visually asymmetric ``conj'' relation
between ``coffee'' and ``tea'' is overloaded to admit a symmetric relationship),
but it is then non-trivial to distinguish shared modifiers from private ones (e.g.,
in the UD tree at the bottom of \autoref{fig:main-ex}, it is difficult to tell that ``hot'' is private to ``coffee'' and ``tea'', which share it,
but ``hot'' does not modify ``bun'').
Another choice is let one of the coordinators dominate the phrase \citep{hajic+01,hajic+20},
but the coordinator does not directly capture
the syntactic category of the coordinated phrase.
Decisions on which of these dependency-based fixes is more workable
are further complicated by the interaction
between representation styles and their learnability in statistical parsing
\citep{nilsson+06,johansson-nugues07,rehbein+17}.

\paragraph{Enhanced UD}
A tactic used by many recent releases of UD treebanks
is to introduce certain extra edges and non-lexical nodes
\citep{schuster-manning16,nivre+18,bouma+20}.
While some of the theoretical issues still persist in this approach
with respect to capturing the symmetric nature of relations between conjuncts,
this solution better represents shared modifiers in coordinations,
and so is a promising direction.
In work concurrent with our own, \citet{grunewald+21} manually correct
the coordination structure annotations in
an English treebank under the enhanced UD representation format.
We leave it to future work to explore the feasibility of
automatic conversion of coordination structure representations
between enhanced UD trees and {\em bubble trees}, which we discuss next.

\subsection{Bubble Trees}
\label{sec:bubble}

An alternative solution to the coordination-in-dependency-trees dilemma
is to permit certain restricted phrase-inspired constructs for such structures.
Indeed, \posscite{tesniere59} seminal work on dependency grammar
does not describe all syntactic relations in terms of dependencies,
but rather reserves a primitive relation for connecting coordinated items.
\citet{hudson84} further extends this idea
by introducing explicit markings of coordination boundaries.

In this paper, we revisit {\em bubble trees},
a representational device along the same vein introduced by \citet{kahane97} for
syntactic representation. (Kahane credits \citet{gladkij68} with a formal study.)
{\em Bubbles} are used to denote coordinated phrases; otherwise, asymmetric
dependency relations are retained.
Conjuncts immediately within the bubble
may co-head the bubble,
and the bubble itself
may establish dependencies with its
governor and modifiers.
\autoref{fig:main-ex} depicts an example bubble tree.

We now formally define bubble trees
and their projective subset,
which will become the focus of our transition-based parser in \autoref{sec:transition}.
The following formal descriptions are
adapted from \citet{kahane97},
tailored to the presentation of our parser.

\paragraph{Formal Definition}
Given a dependency-relation label set $L$,
we define a bubble tree for a length-$n$ sentence $W=w_1,\ldots,w_n$
to be a quadruple $(V,\bubbleset,\phi,A)$,
where $V=\{\rootsym,w_1,\ldots,w_n\}$ is the ground set of nodes
(\rootsym\ is the dummy root),
$\bubbleset$ is a set of bubbles,
the function $\phi: \bubbleset \mapsto
(2^{V}\backslash\{\varnothing\})$ gives the content
of each bubble as a non-empty\footnote{
Our definition does not allow empty nodes;
we leave it to future work to support them
for gapping constructions.
}
subset of $V$, and
$A\subset \bubbleset \times L \times
\bubbleset$ defines
a labeled directed tree over $\bubbleset$.
Given labeled directed tree $A$,
we say $\alpha_1\rightarrow\alpha_2$ if and only if
$(\alpha_1, l, \alpha_2)\in A$ for some $l$.
We denote the reflexive transitive closure of
relation $\rightarrow$
by $\overset{*}{\rightarrow}$.

Bubble tree $(V,\bubbleset,\phi,A)$ is \emph{well-formed} if and only
if it
satisfies the following conditions:\footnote{
We do not use
$\beta$ for bubbles
because we reserve the $\beta$ symbol for our
parser's \underline{b}uffer.}

\noindent $\bullet$
\cond{no partial overlap}
$\forall\alpha_1,\alpha_2\in\bubbleset$,
either $\phi(\alpha_1)\cap\phi
(\alpha_2)=\varnothing$
or $\phi(\alpha_1)\subseteq\phi(\alpha_2)$
or $\phi(\alpha_2)\subseteq\phi(\alpha_1)$;

\noindent
$\bullet$ \cond{non-duplication} there exists no
non-identical $\alpha_1,\alpha_2\in\bubbleset$
such that $\phi(\alpha_1)=\phi(\alpha_2)$;

\noindent
$\bullet$ \cond{lexical coverage} for any singleton (i.e., one-element) set $s$ in $2^{V}$,
$\exists \alpha \in \bubbleset$ such that $\phi(\alpha)=s$;

\noindent
$\bullet$ \cond{roothood} the root \rootsym\
appears in exactly one bubble,
a
singleton
that
is the root of the tree defined by $A$.

\noindent
$\bullet$ \cond{containment}
if $\exists \alpha_1, \alpha_2 \in \bubbleset$ such that
 $\phi(\alpha_2)\subset\phi(\alpha_1)$,
then $\alpha_1\overset{*}{\rightarrow}\alpha_2$.

\paragraph{Projectivity}
Our parser focuses on the subclass of {\em projective}
well-formed bubble trees.
Visually, a projective bubble tree only contains bubbles covering a consecutive sequence of words
(such that we can draw boxes around the span
of words to represent them)
and can be drawn with all arcs arranged spatially above the sentence
where no two arcs or bubble boundaries
cross each other.
The bubble tree in \autoref{fig:main-ex} is projective.

Formally,
we define the projection $\psi(\alpha)\in 2^{V}$ of a bubble $\alpha\in\bubbleset$ to be
all nodes the bubble and its subtree cover,
that is,
$v\in\psi(\alpha)$ if and only if
$\alpha\overset{*}{\rightarrow}\alpha'$
and $v\in\phi(\alpha')$ for some $\alpha'$.
Then, we can define a well-formed bubble tree to be {\em projective} if and only if
it additionally satisfies the following: \par
{ %
\newcommand{\word}{w\xspace}
\noindent
$\bullet$
\cond{continuous coverage}
for any bubble $\alpha\in\bubbleset$,
if $\word_i,\word_j\in\phi(\alpha)$ and
$i<k<j$,
then $\word_k\in\phi(\alpha)$;
} %
{ %
\newcommand{\word}{w\xspace}
\noindent
$\bullet$
\cond{continuous projections}
for any bubble $\alpha\in\bubbleset$,
if $\word_i,\word_j\in\psi(\alpha)$ and
$i<k<j$,
then $\word_k\in\psi(\alpha)$;
} %

\noindent
$\bullet$
\cond{contained projections}
for $\alpha_1,\alpha_2\in\bubbleset$,
if $\alpha_1\overset{*}{\rightarrow}\alpha_2$,
then either $\psi(\alpha_2)\subset\phi(\alpha_1)$
or $\psi(\alpha_2)\cap\phi(\alpha_1)=\varnothing$.

\section{Our Transition System for Parsing Bubble Trees}
\label{sec:transition}

\begin{table*}[!h]
\small
\centering
\begin{tabular}{p{3cm}rlrl}
\toprule
Transition
& \multicolumn{2}{c}{\hspace{40pt}From}
& \multicolumn{2}{c}{\hspace{15pt}To\hfill{}}\\
\quad (Pre-conditions)& Stack \stack & Buffer \buffer & Stack \stack$'$ & Buffer \buffer$'$ \\
\midrule

\shift \newline
\makebox[0pt][l]{\quad($|\buffer| \geq 1$)}
&

\begin{tikzpicture}[baseline=(s0.base)]
\node (s0) {\ldots};
\node (sl) at ($ (s0.base west) + (-0.1, -0.15) $) {};
\node (sr) at ($ (s0.base east) + (0, -0.15) $) {};
\node (sm) at ($ (s0.base east) + (0, 0.05) $) {};
\draw (sl.center) -- (sr.center);
\draw (sr.center) -- (sm.center);
\end{tikzpicture}
&

\begin{tikzpicture}[baseline=(b0.base)]
\node (b0) {\bufferelement{1}};
\node [base right=-5pt of b0] (buffer) {\ldots};
\node (bl) at ($ (b0.base west) + (0., -0.15) $) {};
\node (br) at ($ (b0.base east) + (0.4, -0.15) $) {};
\node (bm) at ($ (b0.base west) + (0, 0.05) $) {};
\draw (bm.center) -- (bl.center);
\draw (bl.center) -- (br.center);
\end{tikzpicture}
&

\begin{tikzpicture}[baseline=(s0.base)]
\node (stack) {\ldots};
\node [base right=-5pt of stack] (s0) {\bufferelement{1}};
\node (sl) at ($ (s0.base west) + (-0.5, -0.15) $) {};
\node (sr) at ($ (s0.base east) + (0., -0.15) $) {};
\node (sm) at ($ (s0.base east) + (0, 0.05) $) {};
\draw (sl.center) -- (sr.center);
\draw (sr.center) -- (sm.center);
\end{tikzpicture}
&

\begin{tikzpicture}[baseline=(b0.base)]
\node (b0) {\ldots};
\node (bl) at ($ (b0.base west) + (0., -0.15) $) {};
\node (br) at ($ (b0.base east) + (0.1, -0.15) $) {};
\node (bm) at ($ (b0.base west) + (0, 0.05) $) {};
\draw (bm.center) -- (bl.center);
\draw (bl.center) -- (br.center);
\end{tikzpicture}
\\

\leftarc\textsubscript{lbl}\newline
\makebox[0pt][l]{\quad($|\stack| \geq 1; |\buffer| \geq 1; \stackelement{1}, \bufferelement{1}\notin \openset; \phi(\stackelement{1}) \neq \{\rootsym\}$)}
&

\begin{tikzpicture}[baseline=(sr.center)]
\node (stack) {\ldots};
\node [base right=-5pt of stack] (s0) {\stackelement{1}};
\node (sl) at ($ (s0.south west) + (-0.5, 0) $) {};
\node (sr) at ($ (s0.south east) $) {};
\node (sm) at ($ (s0.south east) + (0, 0.2) $) {};
\draw (sl.center) -- (sr.center);
\draw (sr.center) -- (sm.center);
\end{tikzpicture}
&

\begin{tikzpicture}[baseline=(bl.center)]
\node (b0) {\bufferelement{1}};
\node [base right=-5pt of b0] (buffer) {\ldots};
\node (bl) at ($ (b0.south west) $) {};
\node (br) at ($ (b0.south east) + (0.4, 0) $) {};
\node (bm) at ($ (b0.south west) + (0, 0.2) $) {};
\draw (bm.center) -- (bl.center);
\draw (bl.center) -- (br.center);
\end{tikzpicture}
&

\begin{tikzpicture}[baseline=(sr.center)]
\node (s0) {\ldots};
\node (sl) at ($ (s0.south west) + (-0.1, 0) $) {};
\node (sr) at ($ (s0.south east) $) {};
\node (sm) at ($ (s0.south east) + (0, 0.2) $) {};
\draw (sl.center) -- (sr.center);
\draw (sr.center) -- (sm.center);
\end{tikzpicture}
&

\begin{tikzpicture}[baseline=(bl.center)]
\node (b0) {\bufferelement{1}};
\node [base right=-5pt of b0] (buffer) {\ldots};
\node [below left=5pt and -10pt of b0] (s0) {\stackelement{1}};
\node (bl) at ($ (b0.south west) $) {};
\node (br) at ($ (b0.south east) + (0.4, 0) $) {};
\node (bm) at ($ (b0.south west) + (0, 0.2) $) {};
\draw (bm.center) -- (bl.center);
\draw (bl.center) -- (br.center);
\draw[edge] (b0.south) -- (s0.north) node [midway, anchor=east, xshift=-3pt] {\scriptsize{lbl}};
\end{tikzpicture}
\\

\rightarc\textsubscript{lbl} \newline
\makebox[0pt][l]{\quad($|\stack| \geq 2; \stackelement{1}, \stackelement{2}\notin \openset$)}
&

\begin{tikzpicture}[baseline=(s1.base)]
\node (stack) {\ldots};
\node [base right=-5pt of stack] (s1) {\stackelement{2}};
\node [base right=-5pt of s1] (s0) {\stackelement{1}};
\node (sl) at ($ (s0.south west) + (-1.0, 0) $) {};
\node (sr) at ($ (s0.south east) $) {};
\node (sm) at ($ (s0.south east) + (0, 0.2) $) {};
\draw (sl.center) -- (sr.center);
\draw (sr.center) -- (sm.center);
\end{tikzpicture}
&

\begin{tikzpicture}[baseline=(b0.base)]
\node (b0) {\ldots};
\node (bl) at ($ (b0.south west) $) {};
\node (br) at ($ (b0.south east) + (0.1, 0) $) {};
\node (bm) at ($ (b0.south west) + (0, 0.2) $) {};
\draw (bm.center) -- (bl.center);
\draw (bl.center) -- (br.center);
\end{tikzpicture}
&

\begin{tikzpicture}[baseline=(stack.base)]
\node (stack) {\ldots};
\node [base right=-5pt of stack] (s0) {\stackelement{2}};
\node [below right=5pt and -10pt of s0] (child) {\stackelement{1}};
\node (sl) at ($ (s0.south west) + (-0.5, 0) $) {};
\node (sr) at ($ (s0.south east) $) {};
\node (sm) at ($ (s0.south east) + (0, 0.2) $) {};
\draw (sl.center) -- (sr.center);
\draw (sr.center) -- (sm.center);
\draw[edge] (s0.south) -- (child.north) node [midway, anchor=west, xshift=3pt] {\scriptsize{lbl}};
\end{tikzpicture}
&

\begin{tikzpicture}[baseline=(b0.base)]
\node (b0) {\ldots};
\node (bl) at ($ (b0.south west) $) {};
\node (br) at ($ (b0.south east) + (0.1, 0) $) {};
\node (bm) at ($ (b0.south west) + (0, 0.2) $) {};
\draw (bm.center) -- (bl.center);
\draw (bl.center) -- (br.center);
\end{tikzpicture}
\\

\bubbleopen\textsubscript{lbl} \newline
\makebox[0pt][l]{\quad($|\stack| \geq 2; \stackelement{1}, \stackelement{2}\notin \openset; \phi(\stackelement{2}) \neq \{\rootsym\}$)}
&

\begin{tikzpicture}[baseline=(sr.center)]
\node (stack) {\ldots};
\node [base right=-5pt of stack] (s1) {\stackelement{2}};
\node [base right=-5pt of s1] (s0) {\stackelement{1}};
\node (sl) at ($ (s0.south west) + (-1.0, 0) $) {};
\node (sr) at ($ (s0.south east) $) {};
\node (sm) at ($ (s0.south east) + (0, 0.2) $) {};
\draw (sl.center) -- (sr.center);
\draw (sr.center) -- (sm.center);
\end{tikzpicture}
&

\begin{tikzpicture}[baseline=(bl.center)]
\node (b0) {\ldots};
\node (bl) at ($ (b0.south west) $) {};
\node (br) at ($ (b0.south east) + (0.1, 0) $) {};
\node (bm) at ($ (b0.south west) + (0, 0.2) $) {};
\draw (bm.center) -- (bl.center);
\draw (bl.center) -- (br.center);
\end{tikzpicture}
&

\begin{tikzpicture}[baseline=(sr.center)]
\node (stack) {\ldots};
\node [base right=-1pt of stack] (s1) {\stackelement{2}};
\node [base right=-1pt of s1] (s0) {\stackelement{1}};
\node (ccp) [draw, densely dotted, rectangle, rounded corners=3pt, fit = {(s0) (s1) ($ (s1.north) + (0, 0.25) $)}, inner sep=0pt] {};
\node (sl) at ($ (ccp.south west) + (-0.6, -0.1) $) {};
\node (sr) at ($ (ccp.south east) + (0.1, -0.1) $) {};
\node (sm) at ($ (ccp.south east) + (0.1, 0.1) $) {};
\draw (sl.center) -- (sr.center);
\draw (sr.center) -- (sm.center);
\node (conjn) at ($ (s1) + (-0.15, 0.65) $) {};
\node (conjs) at ($ (s1) + (-0.15, 0.) $) {};
\draw[edge] (conjn |- ccp.north) -- (conjs) node [midway, anchor=west, xshift=-1.5pt] {\scriptsize{conj}};
\node (conjn2) at ($ (s0) + (-0.1, 0.65) $) {};
\node (conjs2) at ($ (s0) + (-0.1, 0.) $) {};
\draw[edge] (conjn2 |- ccp.north) -- (conjs2) node [midway, anchor=west, xshift=-1pt] {\scriptsize{lbl}};
\end{tikzpicture}
&

\begin{tikzpicture}[baseline=(bl.center)]
\node (b0) {\ldots};
\node (bl) at ($ (b0.south west) $) {};
\node (br) at ($ (b0.south east) + (0.1, 0) $) {};
\node (bm) at ($ (b0.south west) + (0, 0.2) $) {};
\draw (bm.center) -- (bl.center);
\draw (bl.center) -- (br.center);
\end{tikzpicture}
\\

\bubbleattach\textsubscript{lbl} \newline
\makebox[0pt][l]{\quad($|\stack| \geq 2; \stackelement{1}\notin \openset; \stackelement{2}\in \openset$)}
&

\begin{tikzpicture}[baseline=(sr.center)]
\node (stack) {\ldots};
\node [base right=-1pt of stack] (s1) {\stackelement{2a}};
\node [base right=-1pt of s1] (m) {\ldots};
\node (ccp) [draw, densely dotted, rectangle, rounded corners=3pt, fit = {(m) (s1) ($ (s0.north) + (0, 0.25) $)}, inner sep=0pt] {};
\node [right=-1pt of m] (s0) {\stackelement{1}};
\node (sl) at ($ (s0.south west) + (-2.0, -0.1) $) {};
\node (sr) at ($ (s0.south east) + (0, -0.1) $) {};
\node (sm) at ($ (s0.south east) + (0, 0.1) $) {};
\draw (sl.center) -- (sr.center);
\draw (sr.center) -- (sm.center);
\node (conjn) at ($ (s1) + (-0.15, 0.65) $) {};
\node (conjs) at ($ (s1) + (-0.15, 0.) $) {};
\draw[edge] (conjn |- ccp.north) -- (conjs) node [midway, anchor=west, xshift=-1.5pt] {\scriptsize{conj}};
\node (conjnm) at ($ (m) + (-0.15, 0.65) $) {};
\node (conjsm) at ($ (m) + (-0.15, 0.) $) {};
\draw[edge] (conjnm |- ccp.north) -- (conjsm) node [midway, anchor=west, xshift=-1.5pt] {\scriptsize{\ldots}};
\end{tikzpicture}
&

\begin{tikzpicture}[baseline=(bl.center)]
\node (b0) {\ldots};
\node (bl) at ($ (b0.south west) $) {};
\node (br) at ($ (b0.south east) + (0.1, 0) $) {};
\node (bm) at ($ (b0.south west) + (0, 0.2) $) {};
\draw (bm.center) -- (bl.center);
\draw (bl.center) -- (br.center);
\end{tikzpicture}
&

\begin{tikzpicture}[baseline=(sr.center)]
\node (stack) {\ldots};
\node [base right=-1pt of stack] (s1) {\stackelement{2a}};
\node [base right=-1pt of s1] (m) {\ldots};
\node [base right=-1pt of m] (s0) {\stackelement{1}};
\node (ccp) [draw, densely dotted, rectangle, rounded corners=3pt, fit = {(s0) (s1) ($ (s0.north) + (0, 0.25) $)}, inner sep=0pt] {};
\node (sl) at ($ (ccp.south west) + (-0.6, -0.1) $) {};
\node (sr) at ($ (ccp.south east) + (0.1, -0.1) $) {};
\node (sm) at ($ (ccp.south east) + (0.1, 0.1) $) {};
\draw (sl.center) -- (sr.center);
\draw (sr.center) -- (sm.center);
\node (conjn) at ($ (s1) + (-0.15, 0.65) $) {};
\node (conjs) at ($ (s1) + (-0.15, 0.) $) {};
\draw[edge] (conjn |- ccp.north) -- (conjs) node [midway, anchor=west, xshift=-1.5pt] {\scriptsize{conj}};
\node (conjnm) at ($ (m) + (-0.15, 0.65) $) {};
\node (conjsm) at ($ (m) + (-0.15, 0.) $) {};
\draw[edge] (conjnm |- ccp.north) -- (conjsm) node [midway, anchor=west, xshift=-1.5pt] {\scriptsize{\ldots}};
\node (conjn2) at ($ (s0) + (-0.1, 0.65) $) {};
\node (conjs2) at ($ (s0) + (-0.1, 0.) $) {};
\draw[edge] (conjn2 |- ccp.north) -- (conjs2) node [midway, anchor=west, xshift=-1pt] {\scriptsize{lbl}};
\end{tikzpicture}
&

\begin{tikzpicture}[baseline=(bl.center)]
\node (b0) {\ldots};
\node (bl) at ($ (b0.south west) $) {};
\node (br) at ($ (b0.south east) + (0.1, 0) $) {};
\node (bm) at ($ (b0.south west) + (0, 0.2) $) {};
\draw (bm.center) -- (bl.center);
\draw (bl.center) -- (br.center);
\end{tikzpicture}
\\

\bubbleclose \newline
\makebox[0pt][l]{\quad($|\stack| \geq 1; \stackelement{1}\in \openset$)}
&

\begin{tikzpicture}[baseline=(sr.center)]
\node (stack) {\ldots};
\node [base right=-1pt of stack] (s1) {\stackelement{1a}};
\node [base right=-1pt of s1] (s0) {\ldots};
\node (ccp) [draw, densely dotted, rectangle, rounded corners=3pt, fit = {(s0) (s1) ($ (s1.north) + (0, 0.25) $)}, inner sep=0pt] {};
\node (sl) at ($ (ccp.south west) + (-0.6, -0.1) $) {};
\node (sr) at ($ (ccp.south east) + (0.1, -0.1) $) {};
\node (sm) at ($ (ccp.south east) + (0.1, 0.1) $) {};
\draw (sl.center) -- (sr.center);
\draw (sr.center) -- (sm.center);
\node (conjn) at ($ (s1) + (-0.15, 0.65) $) {};
\node (conjs) at ($ (s1) + (-0.15, 0.) $) {};
\draw[edge] (conjn |- ccp.north) -- (conjs) node [midway, anchor=west, xshift=-1.5pt] {\scriptsize{conj}};
\node (conjn2) at ($ (s0) + (-0.1, 0.65) $) {};
\node (conjs2) at ($ (s0) + (-0.1, 0.) $) {};
\draw[edge] (conjn2 |- ccp.north) -- (conjs2) node [midway, anchor=west, xshift=-1pt] {\scriptsize{\ldots}};
\end{tikzpicture}
&

\begin{tikzpicture}[baseline=(bl.center)]
\node (b0) {\ldots};
\node (bl) at ($ (b0.south west) $) {};
\node (br) at ($ (b0.south east) + (0.1, 0) $) {};
\node (bm) at ($ (b0.south west) + (0, 0.2) $) {};
\draw (bm.center) -- (bl.center);
\draw (bl.center) -- (br.center);
\end{tikzpicture}
&

\begin{tikzpicture}[baseline=(sr.center)]
\node (s0) {\ldots};
\node (sl) at ($ (s0.south west) + (-0.1, 0) $) {};
\node (sr) at ($ (s0.south east) $) {};
\node (sm) at ($ (s0.south east) + (0, 0.2) $) {};
\draw (sl.center) -- (sr.center);
\draw (sr.center) -- (sm.center);
\end{tikzpicture}
&

\begin{tikzpicture}[baseline=(bl.center)]
\node [base right=-1pt of stack] (s1) {\stackelement{1a}};
\node [base right=-1pt of s1] (s0) {\ldots};
\node [base right=-1pt of s0] (b0) {\ldots};
\node (ccp) [draw, rectangle, line width=0.8pt, rounded corners=3pt, fit = {(s0) (s1) ($ (s1.north) + (0, 0.25) $)}, inner sep=0pt] {};
\node (conjn) at ($ (s1) + (-0.15, 0.65) $) {};
\node (conjs) at ($ (s1) + (-0.15, 0.) $) {};
\draw[edge] (conjn |- ccp.north) -- (conjs) node [midway, anchor=west, xshift=-1.5pt] {\scriptsize{conj}};
\node (conjn2) at ($ (s0) + (-0.1, 0.65) $) {};
\node (conjs2) at ($ (s0) + (-0.1, 0.) $) {};
\draw[edge] (conjn2 |- ccp.north) -- (conjs2) node [midway, anchor=west, xshift=-1pt] {\scriptsize{\ldots}};
\node (bl) at ($ (ccp.south west) + (-0.1, -0.1) $) {};
\node (br) at ($ (ccp.south east) + (0.6, -0.1) $) {};
\node (bm) at ($ (ccp.south west) + (-0.1, 0.1) $) {};
\draw (bm.center) -- (bl.center);
\draw (bl.center) -- (br.center);
\end{tikzpicture}
\\

\bottomrule

\end{tabular}
\caption{
    Illustration of our \transitionsystem transition system.
    We give the pre-conditions for each transition
    and visualizations of the affected stack and buffer items
    comparing the configurations before and after taking that transition.
    \openset denotes the set of currently open bubbles
    and \rootsym is the dummy root symbol.
}
\label{tbl:transition}

\end{table*}

\tikzstyle{innerbubble} = [draw=Maroon, line width=0.8pt]
\tikzstyle{outerbubble} = [draw=RoyalBlue, line width=0.8pt]

\begin{figure*}[!h]
\centering
\begin{tabular}{@{\hspace{-6pt}}l@{\hspace{-4pt}}|@{\hspace{-2pt}}l@{\hspace{-6pt}}}
\begin{tabular}{S@{\hspace{-3pt}}l}

&
\hspace{5pt}
{\small \textit{Stack}}
\hfill
{\small \textit{Buffer}}
\hspace{5pt}
\\

{\small \colorbox{black}{\textcolor{white}{(Initial)}}}
&

\begin{tikzpicture}[baseline=(root.base)]
\node (root) {\rootsym};
\node (sl) at ($ (root.base west) + (0., -0.15) $) {};
\node (sr) at ($ (root.base east) + (0., -0.15)$) {};
\node (sm) at ($ (root.base east) + (0., 0.05) $) {};
\draw (sl.center) -- (sr.center);
\draw (sr.center) -- (sm.center);
\end{tikzpicture}

\begin{tikzpicture}[baseline=(bun.base)]
\node (i) {I};
\node [base right=-5pt of i] (prefer) {prefer};
\node [base right=-5pt of prefer] (hot) {hot};
\node [base right=-5pt of hot] (coffee) {coffee};
\node [base right=-5pt of coffee] (or) {or};
\node [base right=-5pt of or] (tea) {tea};
\node [base right=-5pt of tea] (and) {and};
\node [base right=-5pt of and] (a) {a};
\node [base right=-5pt of a] (bun) {bun};
\node (bm) at ($ (i.base west) + (0., 0.05) $) {};
\node (bl) at ($ (i.base west) + (0., -0.15) $) {};
\node (br) at ($ (bun.base east) + (0., -0.15) $) {};
\draw (bm.center) -- (bl.center);
\draw (bl.center) -- (br.center);
\end{tikzpicture}
\\

\taketransition{\shift}
&

\begin{tikzpicture}[baseline=(root.base)]
\node (root) {\rootsym};
\node [base right=-5pt of root] (i) {I};
\node (sl) at ($ (root.base west) + (0., -0.15) $) {};
\node (sr) at ($ (i.base east) + (0., -0.15)$) {};
\node (sm) at ($ (i.base east) + (0., 0.05) $) {};
\draw (sl.center) -- (sr.center);
\draw (sr.center) -- (sm.center);
\end{tikzpicture}

\begin{tikzpicture}[baseline=(bun.base)]
\node (prefer) {prefer};
\node [base right=-5pt of prefer] (hot) {hot};
\node [base right=-5pt of hot] (coffee) {coffee};
\node [base right=-5pt of coffee] (or) {or};
\node [base right=-5pt of or] (tea) {tea};
\node [base right=-5pt of tea] (and) {and};
\node [base right=-5pt of and] (a) {a};
\node [base right=-5pt of a] (bun) {bun};
\node (bm) at ($ (prefer.base west) + (0., 0.05) $) {};
\node (bl) at ($ (prefer.base west) + (0., -0.15) $) {};
\node (br) at ($ (bun.base east) + (0., -0.15) $) {};
\draw (bm.center) -- (bl.center);
\draw (bl.center) -- (br.center);
\end{tikzpicture}
\\

\taketransition{\leftarc\textsubscript{nsubj}}
&

\begin{tikzpicture}[baseline=(root.base)]
\node (root) {\rootsym};
\node (sl) at ($ (root.base west) + (0., -0.15) $) {};
\node (sr) at ($ (root.base east) + (0., -0.15)$) {};
\node (sm) at ($ (root.base east) + (0., 0.05) $) {};
\draw (sl.center) -- (sr.center);
\draw (sr.center) -- (sm.center);
\end{tikzpicture}

\begin{tikzpicture}[baseline=(bun.base)]
\node (prefer) {prefer};
\node [below left=5pt and -10pt of prefer] (i) {I};
\node [base right=-5pt of prefer] (hot) {hot};
\node [base right=-5pt of hot] (coffee) {coffee};
\node [base right=-5pt of coffee] (or) {or};
\node [base right=-5pt of or] (tea) {tea};
\node [base right=-5pt of tea] (and) {and};
\node [base right=-5pt of and] (a) {a};
\node [base right=-5pt of a] (bun) {bun};
\node (bm) at ($ (prefer.base west) + (0., 0.05) $) {};
\node (bl) at ($ (prefer.base west) + (0., -0.15) $) {};
\node (br) at ($ (bun.base east) + (0., -0.15) $) {};
\draw (bm.center) -- (bl.center);
\draw (bl.center) -- (br.center);
\draw[edge] ($ (prefer.south) + (0, 0.06) $) -- (i.north) node [midway, anchor=east, xshift=-3pt, yshift=-2pt] {\scriptsize{nsubj}};
\end{tikzpicture}
\\

\taketransition{\shift}
&

\begin{tikzpicture}[baseline=(root.base)]
\node (root) {\rootsym};
\node [base right=-5pt of root] (prefer) {prefer};
\node (sl) at ($ (root.base west) + (0., -0.15) $) {};
\node (sr) at ($ (prefer.base east) + (0., -0.15)$) {};
\node (sm) at ($ (prefer.base east) + (0., 0.05) $) {};
\draw (sl.center) -- (sr.center);
\draw (sr.center) -- (sm.center);
\end{tikzpicture}

\begin{tikzpicture}[baseline=(bun.base)]
\node (hot) {hot};
\node [base right=-5pt of hot] (coffee) {coffee};
\node [base right=-5pt of coffee] (or) {or};
\node [base right=-5pt of or] (tea) {tea};
\node [base right=-5pt of tea] (and) {and};
\node [base right=-5pt of and] (a) {a};
\node [base right=-5pt of a] (bun) {bun};
\node (bm) at ($ (hot.base west) + (0., 0.05) $) {};
\node (bl) at ($ (hot.base west) + (0., -0.15) $) {};
\node (br) at ($ (bun.base east) + (0., -0.15) $) {};
\draw (bm.center) -- (bl.center);
\draw (bl.center) -- (br.center);
\end{tikzpicture}
\\

\taketransition{\shift}
&

\begin{tikzpicture}[baseline=(root.base)]
\node (root) {\rootsym};
\node [base right=-5pt of root] (prefer) {prefer};
\node [base right=-5pt of prefer] (hot) {hot};
\node (sl) at ($ (root.base west) + (0., -0.15) $) {};
\node (sr) at ($ (hot.base east) + (0., -0.15)$) {};
\node (sm) at ($ (hot.base east) + (0., 0.05) $) {};
\draw (sl.center) -- (sr.center);
\draw (sr.center) -- (sm.center);
\end{tikzpicture}

\begin{tikzpicture}[baseline=(bun.base)]
\node (coffee) {coffee};
\node [base right=-5pt of coffee] (or) {or};
\node [base right=-5pt of or] (tea) {tea};
\node [base right=-5pt of tea] (and) {and};
\node [base right=-5pt of and] (a) {a};
\node [base right=-5pt of a] (bun) {bun};
\node (bm) at ($ (coffee.base west) + (0., 0.05) $) {};
\node (bl) at ($ (coffee.base west) + (0., -0.15) $) {};
\node (br) at ($ (bun.base east) + (0., -0.15) $) {};
\draw (bm.center) -- (bl.center);
\draw (bl.center) -- (br.center);
\end{tikzpicture}
\\

\taketransition{\shift}
&

\begin{tikzpicture}[baseline=(root.base)]
\node (root) {\rootsym};
\node [base right=-5pt of root] (prefer) {prefer};
\node [base right=-5pt of prefer] (hot) {hot};
\node [base right=-5pt of hot] (coffee) {coffee};
\node (sl) at ($ (root.base west) + (0., -0.15) $) {};
\node (sr) at ($ (coffee.base east) + (0., -0.15)$) {};
\node (sm) at ($ (coffee.base east) + (0., 0.05) $) {};
\draw (sl.center) -- (sr.center);
\draw (sr.center) -- (sm.center);
\end{tikzpicture}

\begin{tikzpicture}[baseline=(bun.base)]
\node (or) {or};
\node [base right=-5pt of or] (tea) {tea};
\node [base right=-5pt of tea] (and) {and};
\node [base right=-5pt of and] (a) {a};
\node [base right=-5pt of a] (bun) {bun};
\node (bm) at ($ (or.base west) + (0., 0.05) $) {};
\node (bl) at ($ (or.base west) + (0., -0.15) $) {};
\node (br) at ($ (bun.base east) + (0., -0.15) $) {};
\draw (bm.center) -- (bl.center);
\draw (bl.center) -- (br.center);
\end{tikzpicture}
\\

\taketransition{\shift}
&

\begin{tikzpicture}[baseline=(root.base)]
\node (root) {\rootsym};
\node [base right=-5pt of root] (prefer) {prefer};
\node [base right=-5pt of prefer] (hot) {hot};
\node [base right=-5pt of hot] (coffee) {coffee};
\node [base right=-5pt of coffee] (or) {or};
\node (sl) at ($ (root.base west) + (0., -0.15) $) {};
\node (sr) at ($ (or.base east) + (0., -0.15)$) {};
\node (sm) at ($ (or.base east) + (0., 0.05) $) {};
\draw (sl.center) -- (sr.center);
\draw (sr.center) -- (sm.center);
\end{tikzpicture}

\begin{tikzpicture}[baseline=(bun.base)]
\node (tea) {tea};
\node [base right=-5pt of tea] (and) {and};
\node [base right=-5pt of and] (a) {a};
\node [base right=-5pt of a] (bun) {bun};
\node (bm) at ($ (tea.base west) + (0., 0.05) $) {};
\node (bl) at ($ (tea.base west) + (0., -0.15) $) {};
\node (br) at ($ (bun.base east) + (0., -0.15) $) {};
\draw (bm.center) -- (bl.center);
\draw (bl.center) -- (br.center);
\end{tikzpicture}
\\

\taketransition{\bubbleopen\textsubscript{cc}}
&

\begin{tikzpicture}[baseline=(root.base)]
\node (root) {\rootsym};
\node [base right=-5pt of root] (prefer) {prefer};
\node [base right=-5pt of prefer] (hot) {hot};
\node [base right=-1pt of hot] (coffee) {coffee};
\node [base right=-5pt of coffee] (or) {or};
\node (ccp) [innerbubble, draw, densely dotted, rectangle, rounded corners=3pt, fit = {(coffee) (or) ($ (or.base) + (0, 0.71) $)}, inner sep=-1.5pt] {};
\node (sl) at ($ (root.base west) + (0., -0.15) $) {};
\node (sr) at ($ (or.base east) + (0., -0.15)$) {};
\node (sm) at ($ (or.base east) + (0., 0.05) $) {};
\draw (sl.center) -- (sr.center);
\draw (sr.center) -- (sm.center);
\node (conjn) at ($ (coffee.base) + (-0.2, 0.8) $) {};
\node (conjs) at ($ (coffee.base) + (-0.2, 0.18) $) {};
\draw[edge] (conjn |- ccp.north) -- (conjs) node [midway, anchor=west, xshift=-1pt] {\scriptsize{conj}};
\node (conjn2) at ($ (or.base) + (-0.14, 0.8) $) {};
\node (conjs2) at ($ (or.base) + (-0.14, 0.18) $) {};
\draw[edge] (conjn2 |- ccp.north) -- (conjs2) node [midway, anchor=west, xshift=-1pt] {\scriptsize{cc}};

\end{tikzpicture}

\begin{tikzpicture}[baseline=(bun.base)]
\node (tea) {tea};
\node [base right=-5pt of tea] (and) {and};
\node [base right=-5pt of and] (a) {a};
\node [base right=-5pt of a] (bun) {bun};
\node (bm) at ($ (tea.base west) + (0., 0.05) $) {};
\node (bl) at ($ (tea.base west) + (0., -0.15) $) {};
\node (br) at ($ (bun.base east) + (0., -0.15) $) {};
\draw (bm.center) -- (bl.center);
\draw (bl.center) -- (br.center);
\end{tikzpicture}
\\

\taketransition{\shift}
&

\begin{tikzpicture}[baseline=(root.base)]
\node (root) {\rootsym};
\node [base right=-5pt of root] (prefer) {prefer};
\node [base right=-5pt of prefer] (hot) {hot};
\node [base right=-3pt of hot] (coffee) {coffee};
\node [base right=-5pt of coffee] (or) {or};
\node [base right=-3pt of or] (tea) {tea};
\node (ccp) [innerbubble, draw, densely dotted, rectangle, rounded corners=3pt, fit = {(coffee) (or) ($ (or.base) + (0, 0.71) $)}, inner sep=-1.5pt] {};
\node (sl) at ($ (root.base west) + (0., -0.15) $) {};
\node (sr) at ($ (tea.base east) + (0., -0.15)$) {};
\node (sm) at ($ (tea.base east) + (0., 0.05) $) {};
\draw (sl.center) -- (sr.center);
\draw (sr.center) -- (sm.center);
\node (conjn) at ($ (coffee.base) + (-0.2, 0.8) $) {};
\node (conjs) at ($ (coffee.base) + (-0.2, 0.18) $) {};
\draw[edge] (conjn |- ccp.north) -- (conjs) node [midway, anchor=west, xshift=-1pt] {\scriptsize{conj}};
\node (conjn2) at ($ (or.base) + (-0.14, 0.8) $) {};
\node (conjs2) at ($ (or.base) + (-0.14, 0.18) $) {};
\draw[edge] (conjn2 |- ccp.north) -- (conjs2) node [midway, anchor=west, xshift=-1pt] {\scriptsize{cc}};

\end{tikzpicture}

\begin{tikzpicture}[baseline=(bun.base)]
\node (and) {and};
\node [base right=-5pt of and] (a) {a};
\node [base right=-5pt of a] (bun) {bun};
\node (bm) at ($ (and.base west) + (0., 0.05) $) {};
\node (bl) at ($ (and.base west) + (0., -0.15) $) {};
\node (br) at ($ (bun.base east) + (0., -0.15) $) {};
\draw (bm.center) -- (bl.center);
\draw (bl.center) -- (br.center);
\end{tikzpicture}
\\

\taketransition{\bubbleattach\textsubscript{conj}}
&

\begin{tikzpicture}[baseline=(root.base)]
\node (root) {\rootsym};
\node [base right=-5pt of root] (prefer) {prefer};
\node [base right=-5pt of prefer] (hot) {hot};
\node [base right=-3pt of hot] (coffee) {coffee};
\node [base right=-5pt of coffee] (or) {or};
\node [base right=-5pt of or] (tea) {tea};
\node (ccp) [innerbubble, draw, densely dotted, rectangle, rounded corners=3pt, fit = {(coffee) (or) (tea) ($ (or.base) + (0, 0.71) $)}, inner sep=-1.5pt] {};
\node (sl) at ($ (root.base west) + (0., -0.15) $) {};
\node (sr) at ($ (tea.base east) + (0., -0.15)$) {};
\node (sm) at ($ (tea.base east) + (0., 0.05) $) {};
\draw (sl.center) -- (sr.center);
\draw (sr.center) -- (sm.center);
\node (conjn) at ($ (coffee.base) + (-0.2, 0.8) $) {};
\node (conjs) at ($ (coffee.base) + (-0.2, 0.18) $) {};
\draw[edge] (conjn |- ccp.north) -- (conjs) node [midway, anchor=west, xshift=-1pt] {\scriptsize{conj}};
\node (conjn2) at ($ (or.base) + (-0.14, 0.8) $) {};
\node (conjs2) at ($ (or.base) + (-0.14, 0.18) $) {};
\draw[edge] (conjn2 |- ccp.north) -- (conjs2) node [midway, anchor=west, xshift=-2pt] {\scriptsize{cc}};
\node (conjn3) at ($ (tea.base) + (-0.22, 0.8) $) {};
\node (conjs3) at ($ (tea.base) + (-0.22, 0.18) $) {};
\draw[edge] (conjn3 |- ccp.north) -- (conjs3) node [midway, anchor=west, xshift=-2.3pt] {\scriptsize{conj}};

\end{tikzpicture}

\begin{tikzpicture}[baseline=(bun.base)]
\node (and) {and};
\node [base right=-5pt of and] (a) {a};
\node [base right=-5pt of a] (bun) {bun};
\node (bm) at ($ (and.base west) + (0., 0.05) $) {};
\node (bl) at ($ (and.base west) + (0., -0.15) $) {};
\node (br) at ($ (bun.base east) + (0., -0.15) $) {};
\draw (bm.center) -- (bl.center);
\draw (bl.center) -- (br.center);
\end{tikzpicture}
\\

\taketransition{\bubbleclose}
&

\begin{tikzpicture}[baseline=(root.base)]
\node (root) {\rootsym};
\node [base right=-5pt of root] (prefer) {prefer};
\node [base right=-5pt of prefer] (hot) {hot};
\node (sl) at ($ (root.base west) + (0., -0.15) $) {};
\node (sr) at ($ (hot.base east) + (0., -0.15)$) {};
\node (sm) at ($ (hot.base east) + (0., 0.05) $) {};
\draw (sl.center) -- (sr.center);
\draw (sr.center) -- (sm.center);
\end{tikzpicture}

\begin{tikzpicture}[baseline=(bun.base)]
\node (space) {\,};
\node (ccp) [innerbubble, draw, rectangle, line width=0.8pt, rounded corners=3pt, fit = {(space)}, inner sep=0pt, xshift=2pt, yshift=3pt] {};
\node [base right=2pt of space] (and) {and};
\node [base right=-5pt of and] (a) {a};
\node [base right=-5pt of a] (bun) {bun};
\node (bm) at ($ (space.base west) + (0., 0.05) $) {};
\node (bl) at ($ (space.base west) + (0., -0.15) $) {};
\node (br) at ($ (bun.base east) + (0., -0.15) $) {};
\draw (bm.center) -- (bl.center);
\draw (bl.center) -- (br.center);
\end{tikzpicture}
\\

\taketransition{\leftarc\textsubscript{amod}}
&

\begin{tikzpicture}[baseline=(root.base)]
\node (root) {\rootsym};
\node [base right=-5pt of root] (prefer) {prefer};
\node (sl) at ($ (root.base west) + (0., -0.15) $) {};
\node (sr) at ($ (prefer.base east) + (0., -0.15)$) {};
\node (sm) at ($ (prefer.base east) + (0., 0.05) $) {};
\draw (sl.center) -- (sr.center);
\draw (sr.center) -- (sm.center);
\end{tikzpicture}

\begin{tikzpicture}[baseline=(bun.base)]
\node (space) {\,};
\node (ccp) [innerbubble, draw, rectangle, line width=0.8pt, rounded corners=3pt, fit = {(space)}, inner sep=0pt, xshift=2pt, yshift=3pt] {};
\node [below left=10pt and -10pt of ccp] (hot) {hot};
\node [base right=2pt of space] (and) {and};
\node [base right=-5pt of and] (a) {a};
\node [base right=-5pt of a] (bun) {bun};
\node (bm) at ($ (space.base west) + (0., 0.05) $) {};
\node (bl) at ($ (space.base west) + (0., -0.15) $) {};
\node (br) at ($ (bun.base east) + (0., -0.15) $) {};
\draw (bm.center) -- (bl.center);
\draw (bl.center) -- (br.center);
\draw[edge] ($ (ccp.south) + (0, -0.1) $) -- (hot.north) node [midway, anchor=east, xshift=-3pt] {\scriptsize{amod}};
\end{tikzpicture}
\\

\end{tabular}
&
\begin{tabular}{S@{\hspace{-3pt}}l}

\taketransition{\shift}
&

\begin{tikzpicture}[baseline=(root.base)]
\node (root) {\rootsym};
\node [base right=-5pt of root] (prefer) {prefer};
\node [base right=-3pt of prefer] (space) {\,};
\node (ccp) [innerbubble, draw, rectangle, line width=0.8pt, rounded corners=3pt, fit = {(space)}, inner sep=0pt, xshift=2pt, yshift=3pt] {};
\node (sl) at ($ (root.base west) + (0., -0.15) $) {};
\node (sr) at ($ (space.base east) + (0.15, -0.15)$) {};
\node (sm) at ($ (space.base east) + (0.15, 0.05) $) {};
\draw (sl.center) -- (sr.center);
\draw (sr.center) -- (sm.center);
\end{tikzpicture}

\begin{tikzpicture}[baseline=(bun.base)]
\node (and) {and};
\node [base right=-5pt of and] (a) {a};
\node [base right=-5pt of a] (bun) {bun};
\node (bm) at ($ (and.base west) + (0., 0.05) $) {};
\node (bl) at ($ (and.base west) + (0., -0.15) $) {};
\node (br) at ($ (bun.base east) + (0., -0.15) $) {};
\draw (bm.center) -- (bl.center);
\draw (bl.center) -- (br.center);
\end{tikzpicture}
\\

\taketransition{\shift}
&

\begin{tikzpicture}[baseline=(root.base)]
\node (root) {\rootsym};
\node [base right=-5pt of root] (prefer) {prefer};
\node [base right=-3pt of prefer] (space) {\,};
\node (ccp) [innerbubble, draw, rectangle, line width=0.8pt, rounded corners=3pt, fit = {(space)}, inner sep=0pt, xshift=2pt, yshift=3pt] {};
\node [base right=1pt of space] (and) {and};
\node (sl) at ($ (root.base west) + (0., -0.15) $) {};
\node (sr) at ($ (and.base east) + (0., -0.15)$) {};
\node (sm) at ($ (and.base east) + (0., 0.05) $) {};
\draw (sl.center) -- (sr.center);
\draw (sr.center) -- (sm.center);
\end{tikzpicture}

\begin{tikzpicture}[baseline=(bun.base)]
\node (a) {a};
\node [base right=-5pt of a] (bun) {bun};
\node (bm) at ($ (a.base west) + (0., 0.05) $) {};
\node (bl) at ($ (a.base west) + (0., -0.15) $) {};
\node (br) at ($ (bun.base east) + (0., -0.15) $) {};
\draw (bm.center) -- (bl.center);
\draw (bl.center) -- (br.center);
\end{tikzpicture}
\\

\taketransition{\bubbleopen\textsubscript{cc}}
&

\begin{tikzpicture}[baseline=(root.base)]
\node (root) {\rootsym};
\node [base right=-5pt of root] (prefer) {prefer};
\node [base right=-3pt of prefer] (space) {\,};
\node (ccp) [innerbubble, draw, rectangle, line width=0.8pt, rounded corners=3pt, fit = {(space)}, inner sep=0pt, xshift=2pt, yshift=3pt] {};
\node (ccp2) [outerbubble, draw, densely dotted, rectangle, rounded corners=3pt, fit = {(space) (and) ($ (and.base east) + (0.07, 0.71) $) ($ (space.west) + (-0.05, 0.) $)}, inner sep=-1.5pt] {};
\node [base right=3pt of space] (and) {and};
\node (sl) at ($ (root.base west) + (0., -0.15) $) {};
\node (sr) at ($ (and.base east) + (0., -0.15)$) {};
\node (sm) at ($ (and.base east) + (0., 0.05) $) {};
\draw (sl.center) -- (sr.center);
\draw (sr.center) -- (sm.center);
\node (conjn) at ($ (space.base) + (-0.05, 0.8) $) {};
\node (conjs) at ($ (space.base) + (-0.05, 0.18) $) {};
\draw[edge] (conjn |- ccp2.north) -- (conjs) node [midway, anchor=west, xshift=-1pt] {\scriptsize{conj}};
\node (conjn2) at ($ (and.base) + (-0.07, 0.8) $) {};
\node (conjs2) at ($ (and.base) + (-0.07, 0.18) $) {};
\draw[edge] (conjn2 |- ccp2.north) -- (conjs2) node [midway, anchor=west, xshift=-1pt] {\scriptsize{cc}};

\end{tikzpicture}

\begin{tikzpicture}[baseline=(bun.base)]
\node (a) {a};
\node [base right=-5pt of a] (bun) {bun};
\node (bm) at ($ (a.base west) + (0., 0.05) $) {};
\node (bl) at ($ (a.base west) + (0., -0.15) $) {};
\node (br) at ($ (bun.base east) + (0., -0.15) $) {};
\draw (bm.center) -- (bl.center);
\draw (bl.center) -- (br.center);
\end{tikzpicture}
\\

\taketransition{\shift}
&

\begin{tikzpicture}[baseline=(root.base)]
\node (root) {\rootsym};
\node [base right=-5pt of root] (prefer) {prefer};
\node [base right=-3pt of prefer] (space) {\,};
\node (ccp) [innerbubble, draw, rectangle, line width=0.8pt, rounded corners=3pt, fit = {(space)}, inner sep=0pt, xshift=2pt, yshift=3pt] {};
\node (ccp2) [outerbubble, draw, densely dotted, rectangle, rounded corners=3pt, fit = {(space) (and) ($ (and.base west) + (0.05, 0.71) $) ($ (space.west) + (-0.05, 0.) $)}, inner sep=-1.5pt] {};
\node [base right=3pt of space] (and) {and};
\node [base right=-3pt of and] (a) {a};
\node (sl) at ($ (root.base west) + (0., -0.15) $) {};
\node (sr) at ($ (a.base east) + (0., -0.15)$) {};
\node (sm) at ($ (a.base east) + (0., 0.05) $) {};
\draw (sl.center) -- (sr.center);
\draw (sr.center) -- (sm.center);
\node (conjn) at ($ (space.base) + (-0.05, 0.8) $) {};
\node (conjs) at ($ (space.base) + (-0.05, 0.18) $) {};
\draw[edge] (conjn |- ccp2.north) -- (conjs) node [midway, anchor=west, xshift=-1pt] {\scriptsize{conj}};
\node (conjn2) at ($ (and.base) + (-0.07, 0.8) $) {};
\node (conjs2) at ($ (and.base) + (-0.07, 0.18) $) {};
\draw[edge] (conjn2 |- ccp2.north) -- (conjs2) node [midway, anchor=west, xshift=-1pt] {\scriptsize{cc}};

\end{tikzpicture}

\begin{tikzpicture}[baseline=(bun.base)]
\node (bun) {bun};
\node (bm) at ($ (bun.base west) + (0., 0.05) $) {};
\node (bl) at ($ (bun.base west) + (0., -0.15) $) {};
\node (br) at ($ (bun.base east) + (0., -0.15) $) {};
\draw (bm.center) -- (bl.center);
\draw (bl.center) -- (br.center);
\end{tikzpicture}
\\

\taketransition{\leftarc\textsubscript{det}}
&

\begin{tikzpicture}[baseline=(root.base)]
\node (root) {\rootsym};
\node [base right=-5pt of root] (prefer) {prefer};
\node [base right=-3pt of prefer] (space) {\,};
\node (ccp) [innerbubble, draw, rectangle, line width=0.8pt, rounded corners=3pt, fit = {(space)}, inner sep=0pt, xshift=2pt, yshift=3pt] {};
\node (ccp2) [outerbubble, draw, densely dotted, rectangle, rounded corners=3pt, fit = {(space) (and) ($ (and.base west) + (0.05, 0.71) $) ($ (space.west) + (-0.05, 0.) $)}, inner sep=-1.5pt] {};
\node [base right=3pt of space] (and) {and};
\node (sl) at ($ (root.base west) + (0., -0.15) $) {};
\node (sr) at ($ (and.base east) + (0., -0.15)$) {};
\node (sm) at ($ (and.base east) + (0., 0.05) $) {};
\draw (sl.center) -- (sr.center);
\draw (sr.center) -- (sm.center);
\node (conjn) at ($ (space.base) + (-0.05, 0.8) $) {};
\node (conjs) at ($ (space.base) + (-0.05, 0.18) $) {};
\draw[edge] (conjn |- ccp2.north) -- (conjs) node [midway, anchor=west, xshift=-1pt] {\scriptsize{conj}};
\node (conjn2) at ($ (and.base) + (-0.08, 0.8) $) {};
\node (conjs2) at ($ (and.base) + (-0.08, 0.18) $) {};
\draw[edge] (conjn2 |- ccp2.north) -- (conjs2) node [midway, anchor=west, xshift=-1pt] {\scriptsize{cc}};

\end{tikzpicture}

\begin{tikzpicture}[baseline=(bun.base)]
\node (bun) {bun};
\node [below left=7pt and -10pt of bun] (a) {a};
\node (bm) at ($ (bun.base west) + (0., 0.05) $) {};
\node (bl) at ($ (bun.base west) + (0., -0.15) $) {};
\node (br) at ($ (bun.base east) + (0., -0.15) $) {};
\draw (bm.center) -- (bl.center);
\draw (bl.center) -- (br.center);
\draw[edge] (bun.south) -- (a.north) node [midway, anchor=east, xshift=-3pt, yshift=-2pt] {\scriptsize{det}};
\end{tikzpicture}
\\

\taketransition{\shift}
&

\begin{tikzpicture}[baseline=(root.base)]
\node (root) {\rootsym};
\node [base right=-5pt of root] (prefer) {prefer};
\node [base right=-3pt of prefer] (space) {\,};
\node (ccp) [innerbubble, draw, rectangle, line width=0.8pt, rounded corners=3pt, fit = {(space)}, inner sep=0pt, xshift=2pt, yshift=3pt] {};
\node (ccp2) [outerbubble, draw, densely dotted, rectangle, rounded corners=3pt, fit = {(space) (and) ($ (and.base west) + (0.05, 0.71) $) ($ (space.west) + (-0.05, 0.) $)}, inner sep=-1.5pt] {};
\node [base right=3pt of space] (and) {and};
\node [base right=-3pt of and] (bun) {bun};
\node (sl) at ($ (root.base west) + (0., -0.15) $) {};
\node (sr) at ($ (bun.base east) + (0., -0.15)$) {};
\node (sm) at ($ (bun.base east) + (0., 0.05) $) {};
\draw (sl.center) -- (sr.center);
\draw (sr.center) -- (sm.center);
\node (conjn) at ($ (space.base) + (-0.05, 0.8) $) {};
\node (conjs) at ($ (space.base) + (-0.05, 0.18) $) {};
\draw[edge] (conjn |- ccp2.north) -- (conjs) node [midway, anchor=west, xshift=-1pt] {\scriptsize{conj}};
\node (conjn2) at ($ (and.base) + (-0.08, 0.8) $) {};
\node (conjs2) at ($ (and.base) + (-0.08, 0.18) $) {};
\draw[edge] (conjn2 |- ccp2.north) -- (conjs2) node [midway, anchor=west, xshift=-1pt] {\scriptsize{cc}};

\end{tikzpicture}

\begin{tikzpicture}[baseline=(empty.base)]
\node (empty) {\emptybuffer};
\node (bm) at ($ (empty.base west) + (0., 0.05) $) {};
\node (bl) at ($ (empty.base west) + (0., -0.15) $) {};
\node (br) at ($ (empty.base east) + (0., -0.15) $) {};
\draw (bm.center) -- (bl.center);
\draw (bl.center) -- (br.center);
\end{tikzpicture}
\\

\taketransition{\bubbleattach\textsubscript{conj}}
&

\begin{tikzpicture}[baseline=(root.base)]
\node (root) {\rootsym};
\node [base right=-5pt of root] (prefer) {prefer};
\node [base right=-3pt of prefer] (space) {\,};
\node (ccp) [innerbubble, draw, rectangle, line width=0.8pt, rounded corners=3pt, fit = {(space)}, inner sep=0pt, xshift=2pt, yshift=3pt] {};
\node [base right=3pt of space] (and) {and};
\node [base right=-5pt of and] (bun) {bun};
\node (ccp2) [outerbubble, draw, densely dotted, rectangle, rounded corners=3pt, fit = {(space) (bun) ($ (bun.base west) + (0.05, 0.71) $) ($ (space.west) + (-0.05, 0.) $)}, inner sep=-1.5pt] {};
\node (sl) at ($ (root.base west) + (0., -0.15) $) {};
\node (sr) at ($ (bun.base east) + (0., -0.15)$) {};
\node (sm) at ($ (bun.base east) + (0., 0.05) $) {};
\draw (sl.center) -- (sr.center);
\draw (sr.center) -- (sm.center);
\node (conjn) at ($ (space.base) + (-0.05, 0.8) $) {};
\node (conjs) at ($ (space.base) + (-0.05, 0.18) $) {};
\draw[edge] (conjn |- ccp2.north) -- (conjs) node [midway, anchor=west, xshift=-1pt] {\scriptsize{conj}};
\node (conjn2) at ($ (and.base) + (-0.08, 0.8) $) {};
\node (conjs2) at ($ (and.base) + (-0.08, 0.18) $) {};
\draw[edge] (conjn2 |- ccp2.north) -- (conjs2) node [midway, anchor=west, xshift=-1pt] {\scriptsize{cc}};
\node (conjn3) at ($ (bun.base) + (-0.24, 0.8) $) {};
\node (conjs3) at ($ (bun.base) + (-0.24, 0.18) $) {};
\draw[edge] (conjn3 |- ccp2.north) -- (conjs3) node [midway, anchor=west, xshift=-1pt] {\scriptsize{conj}};

\end{tikzpicture}

\begin{tikzpicture}[baseline=(empty.base)]
\node (empty) {\emptybuffer};
\node (bm) at ($ (empty.base west) + (0., 0.05) $) {};
\node (bl) at ($ (empty.base west) + (0., -0.15) $) {};
\node (br) at ($ (empty.base east) + (0., -0.15) $) {};
\draw (bm.center) -- (bl.center);
\draw (bl.center) -- (br.center);
\end{tikzpicture}
\\

\taketransition{\bubbleclose}
&

\begin{tikzpicture}[baseline=(root.base)]
\node (root) {\rootsym};
\node [base right=-5pt of root] (prefer) {prefer};
\node (sl) at ($ (root.base west) + (0., -0.15) $) {};
\node (sr) at ($ (prefer.base east) + (0., -0.15)$) {};
\node (sm) at ($ (prefer.base east) + (0., 0.05) $) {};
\draw (sl.center) -- (sr.center);
\draw (sr.center) -- (sm.center);
\end{tikzpicture}

\begin{tikzpicture}[baseline=(space.base)]
\node (space) {\,};
\node (ccp) [outerbubble, draw, rectangle, line width=0.8pt, rounded corners=3pt, fit = {(space)}, inner sep=0pt, xshift=2pt, yshift=3pt] {};
\node (bm) at ($ (space.base west) + (0., 0.05) $) {};
\node (bl) at ($ (space.base west) + (0., -0.15) $) {};
\node (br) at ($ (space.base east) + (0., -0.15) $) {};
\draw (bm.center) -- (bl.center);
\draw (bl.center) -- (br.center);
\end{tikzpicture}
\\

\taketransition{\shift}
&

\begin{tikzpicture}[baseline=(root.base)]
\node (root) {\rootsym};
\node [base right=-5pt of root] (prefer) {prefer};
\node [base right=-3pt of prefer] (space) {\,};
\node (ccp) [outerbubble, draw, rectangle, line width=0.8pt, rounded corners=3pt, fit = {(space)}, inner sep=0pt, xshift=2pt, yshift=3pt] {};
\node (sl) at ($ (root.base west) + (0., -0.15) $) {};
\node (sr) at ($ (space.base east) + (0.15, -0.15)$) {};
\node (sm) at ($ (space.base east) + (0.15, 0.05) $) {};
\draw (sl.center) -- (sr.center);
\draw (sr.center) -- (sm.center);
\end{tikzpicture}

\begin{tikzpicture}[baseline=(empty.base)]
\node (empty) {\emptybuffer};
\node (bm) at ($ (empty.base west) + (0., 0.05) $) {};
\node (bl) at ($ (empty.base west) + (0., -0.15) $) {};
\node (br) at ($ (empty.base east) + (0., -0.15) $) {};
\draw (bm.center) -- (bl.center);
\draw (bl.center) -- (br.center);
\end{tikzpicture}
\\

\taketransition{\rightarc\textsubscript{obj}}
&

\begin{tikzpicture}[baseline=(root.base)]
\node (root) {\rootsym};
\node [base right=-5pt of root] (prefer) {prefer};
\node [below right=12pt and -10pt of prefer] (space) {\,};
\node (ccp) [outerbubble, draw, rectangle, line width=0.8pt, rounded corners=3pt, fit = {(space)}, inner sep=0pt, xshift=2pt, yshift=3pt] {};
\node (sl) at ($ (root.base west) + (0., -0.15) $) {};
\node (sr) at ($ (prefer.base east) + (0., -0.15)$) {};
\node (sm) at ($ (prefer.base east) + (0., 0.05) $) {};
\draw (sl.center) -- (sr.center);
\draw (sr.center) -- (sm.center);
\draw[edge] ($ (prefer.south) + (0, 0.06) $) -- ($ (space.north) + (0, 0.1) $) node [midway, anchor=west, xshift=3pt, yshift=0pt] {\scriptsize{dobj}};
\end{tikzpicture}

\begin{tikzpicture}[baseline=(empty.base)]
\node (empty) {\emptybuffer};
\node (bm) at ($ (empty.base west) + (0., 0.05) $) {};
\node (bl) at ($ (empty.base west) + (0., -0.15) $) {};
\node (br) at ($ (empty.base east) + (0., -0.15) $) {};
\draw (bm.center) -- (bl.center);
\draw (bl.center) -- (br.center);
\end{tikzpicture}
\\

\taketransition{\rightarc\textsubscript{root}}
&

\begin{tikzpicture}[baseline=(root.base)]
\node (root) {\rootsym};
\node [below right=8pt and -15pt of root] (prefer) {prefer};
\node (sl) at ($ (root.base west) + (0., -0.15) $) {};
\node (sr) at ($ (root.base east) + (0., -0.15)$) {};
\node (sm) at ($ (root.base east) + (0., 0.05) $) {};
\draw (sl.center) -- (sr.center);
\draw (sr.center) -- (sm.center);
\draw[edge] ($ (root.south) + (0, -0.02) $) -- ($ (prefer.north) + (0, 0.) $) node [midway, anchor=west, xshift=3pt, yshift=0pt] {\scriptsize{root}};
\end{tikzpicture}

\begin{tikzpicture}[baseline=(empty.base)]
\node (empty) {\emptybuffer};
\node (bm) at ($ (empty.base west) + (0., 0.05) $) {};
\node (bl) at ($ (empty.base west) + (0., -0.15) $) {};
\node (br) at ($ (empty.base east) + (0., -0.15) $) {};
\draw (bm.center) -- (bl.center);
\draw (bl.center) -- (br.center);
\end{tikzpicture}

\hfill
{\small \colorbox{black}{\textcolor{white}{(Terminal)}}}
\\

\end{tabular}
\\
\end{tabular}
\caption{
    Step-by-step visualization of the stack and buffer 
    during parsing of the example sentence in \autoref{fig:main-ex}.
    For steps following an attachment or \bubbleclose\ transition,
    the detailed subtree or internal bubble structure is omitted
    for visual clarity. For the same reason, we omit drawing the boundaries
    around singleton bubbles.
}
\label{fig:walkthrough}
\end{figure*}
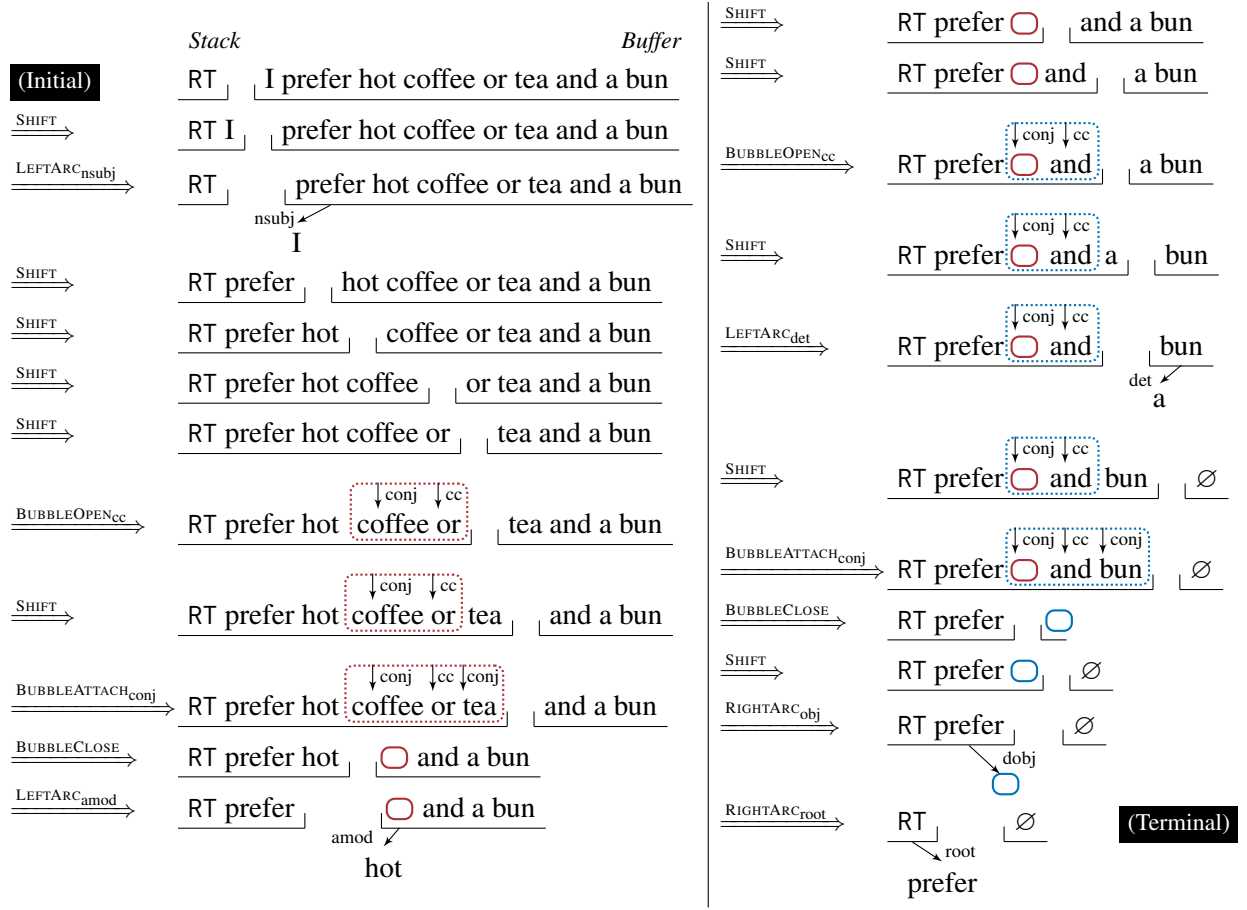

Although, as we have seen, bubble trees
have theoretical benefits in
representing coordination structures
that interface with an overall dependency-based analysis,
there has been a lack of parser implementations
capable of handling such representations.
In this section, we fill this gap
by introducing a transition system that
can incrementally build projective bubble trees.

Transition-based approaches are popular
in dependency parsing \citep{nivre08,kubler+09}.
We propose to extend the \archybrid transition system \citep{kuhlmann+11a}
with transitions specific to bubble structures.\footnote{
Our strategy can be adapted to other transition systems as well;
we focus on \archybrid here
because of its comparatively small inventory of transitions,
absence of spurious ambiguities
(there is a one-to-one mapping between a gold tree and a valid transition sequence),
and abundance of existing implementations \citep[e.g.,][]{kiperwasser-goldberg16a}.
}

\subsection{\transitionsystem Transition System}
\label{sec:bubble-hybrid}

A transition system consists of
a data structure describing the intermediate parser states, called \emph{configurations};
specifications
of the  \emph{initial} and \emph{terminal configurations};
and an inventory of \emph{transitions} that advance the parser in  configuration space
towards reaching a terminal configuration.

Our transition system uses a similar configuration data structure to that of \archybrid,
which consists of a stack, a buffer, and the partially-committed syntactic analysis.
Initially, the stack only contains a singleton bubble corresponding to $\{\rootsym\}$,
and the buffer contains singleton bubbles, each representing a token in the sentence.
Then, through taking transitions one at a time,
the parser can incrementally move items from the buffer to the stack,
or reduce items by attaching them to other bubbles or merging them into larger bubbles.
Eventually, the parser should arrive at a terminal configuration where
the stack contains the singleton bubble of $\{\rootsym\}$ again,
but the buffer is empty
as all the tokens are now attached to or contained in other bubbles
that are now descendants of the $\{\rootsym\}$ singleton,
and we can retrieve a completed bubble-tree parse.

\autoref{tbl:transition} lists the available transitions
in our \transitionsystem system.
The \shift, \leftarc, and \rightarc\ transitions are
as in the \archybrid system.
We  introduce three new transitions to handle coordination-related bubbles:
\bubbleopen\ puts the first two items on the stack into an open bubble,
with the first item in the bubble, i.e., previously the second topmost item on the stack,
labeled as the first conjunct of the resulting bubble;
\bubbleattach\ absorbs the topmost item on the stack
into the open bubble that is at the second topmost position;
and finally, \bubbleclose\ closes the open bubble at the top of the stack
and moves it to the buffer,
which then allows
it
to take modifiers from its left
through \leftarc\ transitions.
\autoref{fig:walkthrough} visualizes the stack and buffer
throughout the process of parsing the example sentence in \autoref{fig:main-ex}.
In particular, the last two steps in the left column of \autoref{fig:walkthrough}
show the  bubble corresponding to the phrase
``coffee or tea'' receiving its left modifier ``hot''
through a \leftarc\ transition
after it is put back on the buffer by a \bubbleclose\ transition.

\paragraph{Formal Definition}
Our transition system is a quadruple $(C,T,c^i,C_\tau)$,
where $C$ is the set of configurations to be defined shortly,
$T$ is the set of transitions
with each element being a partial function $t\in T:C\partialmapsto C$,
$c^i$ maps a sentence to its intial configuration,
and $C_\tau\subset C$ is a set of terminal configurations.
Each configuration $c\in C$ is a septuple
$(\stack, \buffer, V, \bubbleset, \phi, A, \openset)$,
where $V$, $\bubbleset$, $\phi$, and $A$ define a partially-recognized bubble tree,
$\stack$ and $\buffer$ are each an (ordered) list of items in $\bubbleset$,
and $\openset\subset \bubbleset$ is a set of open bubbles.
For a sentence $W=w_1,\ldots, w_n$,
we let $c^i(W)=(\stack^0,\buffer^0,V,\bubbleset^0,\phi^0,\{\},\{\})$,
where $V=\{\rootsym,w_1,\ldots,w_n\}$,
$\bubbleset^0$ contains $n+1$ items,
$\phi^0(\bubbleset^0_0)=\{\rootsym\}$,
$\phi^0(\bubbleset^0_i)=\{w_i\}$ for $i$ from $1$ to $n$,
$\stack^0=[\bubbleset^0_0]$,
and $\buffer^0=[\bubbleset^0_1,\ldots,\bubbleset^0_n]$.
We write $\stack|\stackelement{1}$ and $\bufferelement{1}|\buffer$
to denote a stack and a buffer with their topmost items
being $\stackelement{1}$ and $\bufferelement{1}$
and the remainders being $\stack$ and $\buffer$ respectively.
We also omit the constant $V$ in describing $c$ when the context is clear.

For the transitions $T$, we have:

\noindent
$\bullet$
$\shift[(\sigma,\bufferelement{1}|\buffer,\bubbleset,\phi,A,\openset)]=$

\noindent
\quad$(\sigma|\bufferelement{1},\buffer,\bubbleset,\phi,A,\openset)$;

\noindent
$\bullet$
$\leftarc_\text{lbl}[(\sigma|\stackelement{1},\bufferelement{1}|\buffer,\bubbleset,\phi,A,\openset)]=$

\noindent
\quad$(\sigma,\bufferelement{1}|\buffer,\bubbleset,\phi,A\cup\{(\bufferelement{1},\text{lbl},\stackelement{1})\},\openset)$;

\noindent
$\bullet$
$\rightarc_\text{lbl}[(\sigma|\stackelement{2}|\stackelement{1},\buffer,\bubbleset,\phi,A,\openset)]=$

\noindent
\quad$(\sigma|\stackelement{2},\buffer,\bubbleset,\phi,A\cup\{(\stackelement{2},\text{lbl},\stackelement{1})\},\openset)$;

\noindent
$\bullet$
$\bubbleopen_\text{lbl}[(\sigma|\stackelement{2}|\stackelement{1},\buffer,\bubbleset,\phi,A,\openset)]=$

\noindent
\quad$(\sigma|\alpha,\buffer,\bubbleset\cup\{\alpha\},\phi',A\cup\{(\alpha,\text{conj},\stackelement{2}),\allowbreak(\alpha,\text{lbl},\allowbreak\stackelement{1})\},\openset\cup\{\alpha\})$,
where $\alpha$ is a new bubble,
and
$\phi' = \phi \extendfunction \left\{ \alpha\mapsto\psi(\stackelement{2})\cup\psi(\stackelement{1})\right\}$
(i.e., $\phi'$ is almost the same as $\phi$, but with $\alpha$ added to the function's domain, mapped by the new function to cover the projections of both $\stackelement{2}$ and $\stackelement{1}$); \par
\noindent
$\bullet$
$\bubbleattach_\text{lbl}[(\sigma|\stackelement{2}|\stackelement{1},\buffer,\bubbleset,\phi,A,\openset)]=$

\noindent
\quad$(\sigma|\stackelement{2},\buffer,\bubbleset,\phi',A\cup\{\stackelement{2},\text{lbl},\stackelement{1}\},\openset)$,
where $\phi'=\phi \extendfunction \left\{\stackelement{2}\mapsto\phi(\stackelement{2})\cup\psi(\stackelement{1})\right\}$;

\noindent
$\bullet$
$\bubbleclose[(\sigma|\stackelement{1},\buffer,\bubbleset,\phi,A,\openset)]=$

\noindent
\quad$(\sigma,\stackelement{1}|\buffer,\bubbleset,\phi,A,\openset\backslash\{\stackelement{1}\})$.

\subsection{Soundness and Completeness}
\label{sec:correctness}

In this section,
we show that our
\transitionsystem transition system is both sound and complete (defined below)
with respect to the subclass of projective bubble trees.\footnote{
More precisely,
our transition system handles the subset
where each non-singleton bubble has $\ge 2$ internal children.
}

Define a {\em valid} transition sequence $\pi=t_1,\ldots,t_m$
for a given sentence $W$
to be a sequence
such that for the corresponding sequence of configurations $c_0,\ldots,c_m$,
we have
$c_0=c^i(W)$,
$c_i=t_i(c_{i-1})$,
and $c_m\in C_\tau$,
We can then state soundness and completeness properties,
and present high-level proof sketches below,
adapted from \posscite{nivre08} proof frameworks.

\begin{lemma}
\emph{(Soundness)}
Every valid transition sequence $\pi$ produces a projective bubble tree.
\end{lemma}

\vspace{-1em}
\begin{sproof}
We examine the requirements for a projective bubble tree one by one.
The set of edges satisfies the tree constraints
since every bubble except for the singleton bubble of \rootsym
must have an in-degree of one to have been reduced from the stack,
and the topological order of reductions implies acyclicness.
Lexical coverage is guaranteed by $c^i$.
Roothood is safeguarded by the transition pre-conditions.
Non-duplication is ensured because newly-created bubbles are strictly larger.
All the other properties can be proved by induction over
the lengths of transition sequence prefixes
since each of our transitions preserves zero partial overlap, containment, and projectivity constraints.
\end{sproof}

\begin{lemma}
\emph{(Completeness)}
For every projective bubble tree over any given sentence $W$,
there exists a corresponding valid transition sequence $\pi$.
\end{lemma}

\vspace{-1em}
\begin{sproof}
The proof proceeds by strong induction on sentence length.
We omit relation labels without loss of generality.
The base case of $|W|=1$ is trivial.
For the inductive step,
we enumerate how to decompose the tree's top-level structure.
(1) When the root has multiple children:
Due to projectivity, each child bubble tree $\tau_i$
covers a consecutive span of words
$w_{x_i},\ldots,w_{y_i}$ that are shorter than $|W|$.
Based on the induction hypothesis, there exisits a valid transition sequence $\pi_i$
to construct the child tree over $\rootsym,w_{x_i},\ldots,w_{y_i}$.
Here we let $\pi_i$ to denote the transition sequence excluding the always-present final $\rightarc$ transition
that attaches the subtree to $\rootsym$;
this is for explicit illustration of what transitions to take once the subtrees are constructed.
The full tree can be constructed by $\pi=\pi_1$, \rightarc, $\pi_2$, \rightarc, \ldots (expanding each $\pi_i$ sequence into its component transitions),
where we simply attach each subtree to $\rootsym$ immediately after it is constructed.
(2) When the root has a single child bubble $\alpha$,
we cannot directly use the induction hypothesis since $\alpha$ covers
the same number of words as $W$.
Thus we need to further enumerate the top-level structure of $\alpha$.
(2a) If $\alpha$ has children with their projections outside of $\phi(\alpha)$,
then
we can find a sequence $\pi_0$ for constructing the shorter-length bubble $\alpha$ and placing it on the buffer
(this corresponds to an empty transition sequence if $\alpha$ is a singleton; otherwise, $\pi_0$ ends with a $\bubbleclose$ transition)
and $\pi_i$s for $\alpha$'s outside children; say it has $l$ children left of its contents.
We construct the entire tree via $\pi=\pi_1$,\ldots,$\pi_l$, $\pi_0$, \leftarc, \ldots, \leftarc, \shift,
$\pi_{l+1}$, \rightarc, \ldots, \rightarc,
where we first construct all the left outside children and leave them on the stack,
next build the bubble $\alpha$ and use $\leftarc$ transitions to attach
its left children while it is on the buffer,
then shift $\alpha$ to the stack before
finally continuing on building its right children subtrees,
each immediately followed by a $\rightarc$ transition.
(2b)
If $\alpha$ is a non-singleton bubble without any outside children,
but each of its inside children can be parsed through $\pi_i$ based on the inductive hypothesis,
then we can define $\pi=\pi_1$,$\pi_2$, \bubbleopen, $\pi_3$, \bubbleattach, \ldots, \bubbleclose, \shift, \rightarc,
where we use a $\bubbleopen$ transition once the first two bubble-internal children are built, each subsequent child is attached via $\bubbleattach$ immediately after construction,
and the final three transitions ensure proper closing of the bubble and its attachment to $\rootsym$.
\end{sproof}

\section{Models}
\label{sec:model}

Our model architecture largely follows that of \posscite{kiperwasser-goldberg16a}
neural \archybrid parser,
but we additionally introduce feature composition for non-singleton bubbles,
and a rescoring module  to
reduce frequent coordination-boundary prediction errors.
Our model has five components:
feature extraction, bubble-feature composition,
transition scoring, label scoring, and boundary subtree rescoring.

\paragraph{Feature Extraction}
We first extract contextualized features for each token
using a bidirectional LSTM \citep{graves-schmidhuber05}:
\begin{equation*}
[\wvec{0},\wvec{1},\ldots,\wvec{n}]=\text{bi-LSTM}([\rootsym,w_1,\ldots,w_n]),
\end{equation*}
where the inputs to the bi-LSTM
are concatenations of
word embeddings, POS-tag embeddings, and character-level LSTM embeddings.
We also report experiments  replacing the bi-LSTM
with pre-trained BERT features \citep{devlin+19}.

\paragraph{Bubble-Feature Composition}
We initialize the features\footnote{
We adopt the convenient abuse of notation of allowing indexing by arbitrary objects.
}
for
each singleton bubble $\bubbleset_i$ in the initial configuration
to be $\vvec{\bubbleset_i}=\wvec{i}$.
For a non-singleton bubble $\alpha$,
we use recursively composed features
\begin{equation*}
\vvec{\alpha}=g(\{\vvec{\alpha'}|(\alpha,\text{conj},\alpha')\in A\}),
\end{equation*}
where $g$ is a composition function
combining features from the co-heads (conjuncts) immediately inside the bubble.\footnote{
Comparing with the subtree-feature composition functions in dependency parsing
that are motivated by asymmetric headed constructions \citep{dyer+15a,delhoneux+19,basirat-nivre21},
our definition focuses on composing features
from an unordered set of vectors
representing the conjuncts in a bubble.
The composition function is recursively applied
when there are nested bubbles.
}
For our model, for any $V'= \{\vvec{i_1}, \ldots, \vvec{i_N}\}$,
we set
\begin{equation*}
g(V') =  \tanh(\mathbf{W}^g\cdot\text{mean}(V')),
\end{equation*}
where $\text{mean}()$ computes element-wise averages
and $\mathbf{W}^g$ is
a learnable square matrix.
We also experiment with a parameter-free version:
$g=\text{mean}$.
Neither of the feature functions distinguishes between open and closed bubbles,
so we append to each $\vvec{}$ vector an indicator-feature embedding
based on whether the bubble is open, closed, or singleton.

\paragraph{Transition Scoring}
Given the current parser configuration $c$,
the model predicts the best unlabeled transition
to take among all valid transitions $\text{valid}(c)$
whose pre-conditions are satisfied.
We model the log-linear probability of taking an action
with a multi-layer perceptron (MLP):
\begin{equation*}
P(t|c)\propto \exp(\text{MLP}^\text{trans}_t([\vvec{\stackelement{3}}\circ\vvec{\stackelement{2}}\circ\vvec{\stackelement{1}}\circ\vvec{\bufferelement{1}}])),
\end{equation*}
where $\circ$ denotes vector concatenation, $s_1$ through $s_3$ are the first
through third topmost items on the stack, and $b_1$ is the immediately accessible
buffer item.
We experiment with varying the number of stack items to
extract features from.

\paragraph{Label Scoring}
We separate edge-label prediction from (unlabeled) transition prediction,
but the scoring function takes a similar form:
\begin{equation*}
P(l|c,t)\propto \exp(\text{MLP}^\text{lbl}_l([\vvec{h(c,t)}\circ\vvec{d(c,t)}])),
\end{equation*}
where $(h(c,t),l,d(c,t))$ is the edge to be added into
the partial bubble tree in $t(c)$.

\paragraph{Boundary Subtree Rescoring}
In our preliminary error analysis,
we find that our models tend to make more mistakes
at the boundaries of full coordination phrases
than at the internal conjunct boundaries,
due to incorrect attachments of children
choosing between the phrasal bubble and the first/last conjunct.
For example, our initial model predicts ``if you \underline{\emph{owned it}} and \underline{\emph{liked it Friday}}''
instead of the annotated ``if you \underline{\emph{owned it}} and \underline{\emph{liked it}} Friday''
(the predicted and gold conjuncts are both italicized and underlined),
incorrectly attaching ``Friday'' to ``liked''.
We attribute this problem to the greedy nature of
our first formulation of the parser,
and propose to mitigate the issue through rescoring.
To rescore boundary attachments of a non-singleton bubble $\alpha$,
for each of the left dependents
 $d$ of $\alpha$ and its first conjunct $\alpha_f$,
we (re)-decide the attachment via
\begin{equation*}
P(\alpha\rightarrow d|\alpha_f)=\text{logistic}(\text{MLP}^\text{re}([\vvec{d}\circ\vvec{\alpha}\circ\vvec{\alpha_f}])),
\end{equation*}
and similarly for the last conjunct $\alpha_l$ and a potential right dependent.

\paragraph{Training and Inference}
Our parser is a locally-trained greedy parser.
In training, we optimize the model parameters to maximize the log-likelihoods
of predicting the target transitions and labels along
the paths  generating the gold bubble trees,
and the log-likelihoods of the correct attachments in rescoring;\footnote{
We leave the definition of dynamic oracles \citep{goldberg-nivre13a} for bubble tree parsing to future work.
}
during inference,
the parser greedily commits to the highest-scoring transition and label
for each of its current parser configurations,
and after reaching a terminal configuration,
it rescores and readjusts all boundary subtree attachments.

\section{Empirical Results}
\label{sec:exp}

\begin{table}[t]
    \centering
   \begin{tabular}{lcccc}
    \toprule
     & \multicolumn{2}{c}{Exact} & \multicolumn{2}{c}{Inner} \\
     & All & NP & All & NP \\
    \midrule
    \citetalias{ficler-goldberg16} & -- & -- & \resnumber{72.70} & \resnumber{76.10} \\
    \citetalias{teranishi+17} & \resnumber{71.08} & \resnumber{75.01} & \resnumber{73.74} & \resnumber{77.25} \\
    \citetalias{teranishi+19} & \resnumber{75.47} & \resnumber{77.83} & \resnumber{77.74} & \resnumber{80.06} \\
    Ours       & \resnumber{76.48} & \resnumber{81.63} & \resnumber{78.30} & \resnumber{84.03} \\
    ~~+\bert & \resnumber{83.74} & \resnumber{85.26} & \resnumber{84.46} & \resnumber{86.22} \\
    \bottomrule
   \end{tabular}
   \caption{
       F1 scores on the PTB test set.
       See \autoref{app:results} for precision, recall and dev set results.
   }
   \label{tbl:ptb-main}
\end{table}

\begin{table}[t]
    \centering
   \begin{tabular}{lcc}
    \toprule
     & Exact & Whole \\
    \midrule
    \citetalias{hara+09} & -- & \resnumber{61.5~~} \\
    \citetalias{ficler-goldberg16} & -- & \resnumber{64.14} \\
    \citetalias{teranishi+17} & \resnumber{55.22} & \resnumber{66.31} \\
    \citetalias{teranishi+19} & \resnumber{61.22} & \resnumber{61.31} \\
    Ours       & \resnumber{67.09} & \resnumber{68.23} \\
    ~~+\bert & \resnumber{79.18} & \resnumber{80.41} \\
    \bottomrule
   \end{tabular}
   \caption{
       Recall results on the GENIA dataset
       (we report recall instead of F1 scores following prior work).
       See \autoref{app:results} for detailed results per constituent type.
   }
   \label{tbl:genia-main}
\end{table}

\begin{table}[t]
    \small
    \centering
   \begin{tabular}{lcccc}
    \toprule
     & \multicolumn{2}{c}{Bubble-Hybrid (Ours)} & \multicolumn{2}{c}{Edge-Factored} \\
     & Prec. & Rec. & Prec. & Rec. \\
    \midrule
    punct & \resnumber{92.56} & \resnumber{92.52} & \resnumber{92.92} & \resnumber{92.85} \\
    case  & \resnumber{97.46} & \resnumber{98.14} & \resnumber{97.71} & \resnumber{98.26} \\
    compound & \resnumber{94.12} & \resnumber{95.18} & \resnumber{94.24} & \resnumber{95.02} \\
    det & \resnumber{98.85} & \resnumber{99.13} & \resnumber{98.70} & \resnumber{99.06} \\
    nsubj & \resnumber{97.69} & \resnumber{97.72} & \resnumber{98.01} & \resnumber{97.92} \\
    nmod & \resnumber{93.04} & \resnumber{93.45} & \resnumber{93.20} & \resnumber{93.62} \\
    amod & \resnumber{94.52} & \resnumber{93.43} & \resnumber{94.61} & \resnumber{93.95} \\
    \ldots & \ldots & \ldots & \ldots & \ldots \\
    conj & \resnumber{92.68} & \resnumber{93.20} & \resnumber{92.52} & \resnumber{93.04} \\
    \midrule
    UAS & \multicolumn{2}{c}{\resnumber{95.81}} & \multicolumn{2}{c}{\resnumber{95.99}} \\
    LAS & \multicolumn{2}{c}{\resnumber{94.46}} & \multicolumn{2}{c}{\resnumber{94.56}} \\
    \bottomrule
   \end{tabular}
   \caption{
       PTB test-set results,
       comparing our transition-based bubble parser and an edge-factored graph-based parser, both using a \bert-based feature encoder.
       The relation labels are ordered by decreasing frequency.
       While our transition-based bubble parser slightly underperforms the graph-based dependency parser generally,
       perhaps due to the disadvantage of greedy decoding,
       it gives slightly better precision and recall on the ``conj'' relation type.
   }
   \label{tbl:parsing}
\end{table}

\begin{table}[t]
    \centering
   \begin{tabular}{lcc}
    \toprule
    Rescoring & $+$ & $-$ \\
    \midrule
    Ours (bi-LSTM) & \resnumber{77.10} & \resnumber{76.27} \\
    $\bullet$ $g=\text{mean}$ & \resnumber{75.51} & \resnumber{74.16} \\
    $\bullet$ $\{\vvec{\stackelement{2}},\vvec{\stackelement{1}},\vvec{\bufferelement{1}}\}$ &
    \resnumber{76.05} & \resnumber{74.87} \\
    $\bullet$ $\{\vvec{\stackelement{1}},\vvec{\bufferelement{1}}\}$ &
    \resnumber{76.33} & \resnumber{73.85} \\
    $\bullet$ $\{\vvec{\bufferelement{1}}\}$ &
    \resnumber{50.27} & \resnumber{35.14} \\
    $\bullet$ +\bert & \resnumber{84.40} & \resnumber{83.70} \\
    \bottomrule
   \end{tabular}
   \caption{
       Exact F1 scores of different model variations
       on the PTB dev set,
       w/ and w/o the rescoring module.
   }
   \label{tbl:ablation}
\end{table}

\begin{table}[t]
    \centering
   \begin{tabular}{lccc}
    \toprule
    Complexity & All & Simple & Complex \\
    \midrule
    \citetalias{teranishi+17} & \resnumber{66.09} & \resnumber{72.90} & \resnumber{50.37} \\
    \citetalias{teranishi+19} & \resnumber{70.90} & \resnumber{78.16} & \resnumber{54.16} \\
    Ours & \resnumber{72.97} & \resnumber{79.97} & \resnumber{56.82} \\
    ~~+\bert & \resnumber{80.07} & \resnumber{83.74} & \resnumber{71.59} \\
    \bottomrule
   \end{tabular}
   \caption{
       Per-sentence exact match on the PTB test set.
       \emph{Simple} includes sentences with only one two-conjunct coordination,
       and \emph{complex} contains the other cases.
   }
   \label{tbl:complexity}
\end{table}

\paragraph{Task and Evaluation}
We validate the utility of our transition-based parser
using the task of coordination structure prediction.
Given an input sentence,
the task is to identify all coordination structures
and the spans for all their conjuncts within that sentence.
We mainly evaluate based on \emph{exact} metrics
which count a prediction of a coordination structure as correct
if and only if all of its conjunct spans are correct.
To facilitate comparison with pre-existing systems that
do not attempt to identify all conjunct boundaries,
following \citet{teranishi+17,teranishi+19},
we also consider \emph{inner}
(=only consider the correctness of the two conjuncts adjacent to the coordinator)
and \emph{whole}
(=only consider the boundary of the whole coordinated phrase)
metrics.

\paragraph{Data and Experimental Setup}
We experiment with two English datasets,
the Penn Treebank \citep[PTB;][newswire]{marcus+93} with
added coordination annotations  \citep{ficler-goldberg16b}
and the GENIA treebank \citep[][research abstracts]{kim+03}.
We use the conversion tool distributed with the Stanford Parser \citep{schuster-manning16}
to extract UD trees from the PTB-style phrase-structure annotations,
which we then merge with coordination annotations
to form bubble trees.
We follow prior work in reporting PTB results on its standard splits and
GENIA results using $5$-fold cross-validation.\footnote{We affirm that, as is best practice, only two test-set/crossval-suite runs occurred (one with \bert and one without), happening after we fixed everything else; that is, no other models were tried after seeing the first test-set/cross-validation results with and without \bert.}
During training (but not test), we discard all non-projective sentences.
See \autoref{app:data} for dataset pre-processing and statistics
and \autoref{app:implementation} for implementation details.

\paragraph{Baseline Systems}
We compare our models with several baseline systems.
\citet[HSOM09]{hara+09} use edit graphs to
explicitly align coordinated conjuncts
based on the idea that they are usually similar;
\citet[FG16]{ficler-goldberg16} score candidate
coordinations extracted from a phrase-structure parser
by modeling their symmetry and replaceability properties;
\citet[TSM17]{teranishi+17} directly predict boundaries of coordinated phrases
and then split them into conjuncts;\footnote{
We report results for the extended model of \citetalias{teranishi+17}
as described by \citet{teranishi+19}.
}
\citet[TSM19]{teranishi+19} use separate neural models
to score the inner and outer boundaries of conjuncts
relative to the coordinators,
and then use a chart parser
to find the globally-optimal coordination structures.

\paragraph{Main Results}
\autoref{tbl:ptb-main} and \autoref{tbl:genia-main}
show the main evaluation results on the PTB and GENIA datasets.
Our models
surpass all prior
results on both datasets. While the \bert improvements may not seem surprising,
we note that \citet{teranishi+19} report
that their pre-trained language models --- specifically, static ELMo embeddings ---  fail to improve their model performance.

\paragraph{General Parsing Results}
We also evaluate our models on standard parsing metrics by converting
the predicted bubble trees to UD-style dependency trees.
On PTB, our parsers reach unlabeled and labeled attachment scores (UAS/LAS)
of $95.81$/$94.46$ with \bert and $94.49$/$92.88$ with bi-LSTM,
which are similar to the scores of prior transition-based parsers
equipped with similar feature extractors \citep{kiperwasser-goldberg16a,mohammadshahi-henderson20}.\footnote{
Results are not strictly comparable with
previous PTB evaluations that mostly focus on
non-UD dependency conversions.
\autoref{tbl:parsing} makes a self-contained comparison using the same UD-based and coordination-merged data conversions.
}
\autoref{tbl:parsing} compares the general parsing results of
our bubble parser and an edge-factored graph-based dependency parser
based on \posscite{dozat-manning17} parser architecture and the same feature encoder as our parser and trained on the same data.
Our bubble parser shows a slight improvement on identifying the ``conj'' relations,
despite having a lower overall accuracy due to the greedy nature of our transition-based decoder.
Additionally, our bubble parser simultaneously predicts the boundaries of each coordinated phrase and conjuct,
while a typical dependency parser cannot produce such structures.

\paragraph{Model Analysis}
\autoref{tbl:ablation} shows results of
our models with alternative bubble-feature composition functions
and varying feature-set sizes.
We find that the parameterized form of composition function $g$ performs better,
and the F1 scores mostly
 degrade as we use fewer features from the stack.
Interestingly, the importance of our rescoring module becomes more prominent
when we use fewer features.
Our results resonate with \posscite{shi+17} findings on \archybrid
that we need at least one stack item but not necessarily two.
\autoref{tbl:complexity}
shows that our model
performs better than previous methods on complex sentences
with multiple coordination structures and/or more than two conjuncts,
especially when we use \bert as feature extractor.

\section{Related Work}
\label{sec:related}

\paragraph{Coordination Structure Prediction}

Very early work with heuristic, non-learning-based approaches  \citep{agarwal-boggess92,kurohashi-nagao94a} typically report difficulties
in distinguishing shared modifiers from private ones, although such heuristics have been
recently incorporated in unsupervised work \citep{sawada+20}.
Generally,
researchers have
focused on
symmetry principles, seeking to align conjuncts  \citep{kurohashi-nagao94a,shimbo-hara07,hara+09,hanamoto+12}, since
coordinated conjuncts tend to be
semantically and syntactically similar \citep{hogan07}, as attested to by
psycholinguistic evidence of
structural parallelism \citep{frazier+84,frazier+00,dubey+05}.
\citet{ficler-goldberg16b} and \citet{teranishi+17}
additionally leverage the linguistic principle of replaceability ---
one can typically replace a coordinated phrase
with one of its conjuncts
without the sentence becoming incoherent;
this idea has resulted in improved open information extraction \citep{saha-mausam18}.
Using these principles may further improve our parser.

\paragraph{Coordination in Constituency Grammar}

While our paper mainly focuses on enhancing dependency-based
syntactic analysis with coordination structures,
coordination is a well-studied topic in constituency-based syntax \citep{zhang09},
including proposals and treatments under
lexical functional grammar \citep{kaplan-maxwelliii88},
tree-adjoining grammar \citep{sarkar-joshi96,han-sarkar17a},
and combinatory categorial grammar \citep{steedman96,steedman00}.

\paragraph{Tesni\`ere Dependency Structure}

\citet{sangati-mazza09} propose a representation
that is faithful to \posscite{tesniere59} original framework.
Similar to bubble trees,
their structures include special attention to coordination structures
respecting conjunct symmetry,
but they also include
constructs to handle other syntactic notions currently beyond our parser's scope.\footnote{For example, differentiating content and function words
which has recently been explored by \citet{basirat-nivre21}.}
Such representations have been
used for re-ranking \cite{sangati10},
but not for (direct) parsing.
Perhaps our work can inspire a future Tesni\`ere Dependency Structure parser.

\paragraph{Non-constituent Coordination}

Seemingly incomplete (non-constituent) conjuncts are particularly challenging
\citep{milward94}, and our bubble parser currently has no special mechanism for them.
Dependency-based analyses have adapted by extending to a graph structure \citep{gerdes-kahane15}
or explicitly representing elided elements \citep{schuster+17}.
It may be straightforward to integrate the latter into our parser,
\`a la \posscite{kahane97} proposal of phonologically-empty bubbles.

\section{Conclusion}
\label{sec:conclusion}

We revisit \posscite{kahane97} bubble tree representations
for explicitly encoding coordination boundaries
as a viable alternative to existing mechanisms in dependency-based analysis
of coordination structures.
We introduce a transition system
that is both sound and complete with respect to
the subclass of projective bubble trees.
Empirically, our bubble parsers achieve state-of-the-art results
on the task of coordination structure prediction
on two English datasets.
Future work may extend the research scope to other languages,
graph-based, and non-projective parsing methods.

\paragraph{Acknowledgements}

We thank the anonymous reviewers for their constructive comments,
Yue Guo for discussion,
and Hiroki Teranishi for help with experiment setup.
This work was supported in part by a Bloomberg Data Science Ph.D. Fellowship to Tianze Shi and a gift from Bloomberg to Lillian Lee.

\putbib[ref]
\end{bibunit}

\clearpage
\begin{appendices}

\setcounter{table}{0}
\renewcommand{\thetable}{A\arabic{table}}

\begin{bibunit}[acl_natbib]

\section{Dataset Processing and Statistics}
\label{app:data}

We follow \citet{teranishi+19} and use the same dataset splits and pre-processing steps.
For the Penn Treebank \citep[PTB;][]{marcus+93} data
with added coordination annotations \citep{ficler-goldberg16b},
we use  WSJ sections 02-21 for training, section 22 for development,
and section 23 for test sets respectively.
We also use \posscite{teranishi+19} pre-processing steps
in stripping quotation marks surrounding PTB coordinated phrases
to normalize irregular coordinations.
This results in $39{,}832$/$1{,}700$/$2{,}416$ sentences
and $19{,}890$/$848$/$1{,}099$ coordination structures in train/dev/test splits respectively.
For the GENIA treebank \citep{kim+03},
we use the beta version of the corpus
and follow the same $5$-fold cross-validation splits as \citet{teranishi+19}.
In total, GENIA contains $2{,}508$ sentences and $3{,}598$ coordination structures.

To derive bubble tree representations,
we first convert the PTB-style phrase-structure trees in both treebanks
with the conversion tool \citep{schuster-manning16} provided by the Stanford CoreNLP toolkit version 4.2.0
into Universal Dependencies \citep[UD;][]{nivre+16} style.
We then merge the UD trees with the bubbles formed
by the coordination boundaries.
We define the boundaries to be
from the beginning of the first conjunct to the end of the last conjunct
for each coordinated phrase.
We attach all conjuncts to their corresponding bubbles with a ``conj'' label,
and map any ``conj''-labeled dependencies
outside an annotated coordination to ``dep''.
We resolve modifier scope ambiguities according to conjunct annotations:
if the modifier is within the span of a conjunct,
then it is a private modifier;
otherwise, it is a shared modifier to the entire coordinated phrase
and we attach it to the phrasal bubble.
Since our transition system targets projective bubble trees,
we filter out any non-projective trees during training
(but still evaluate on them during testing).
We retain $39{,}678$ sentences, or $99.6\%$ of the PTB training set,
and $2{,}429$ sentences, or $96.9\%$ of the GENIA dataset.

\section{Implementation Details}
\label{app:implementation}

Our implementation (\url{https://www.github.com/tzshi/bubble-parser-acl21}) is based on PyTorch.

We train our models by using the Adam optimizer \citep{kingma-ba15}.
After a fixed number of optimization steps
($3{,}200$ steps for PTB and $800$ steps for GENIA,
based on their training set sizes),
we perform an evaluation on the dev set.
If the dev set performance fails to improve within $5$ consecutive evaluation rounds,
we multiply the learning rate by $0.1$.
We terminate model training when the learning rate has dropped three times,
and select the best model checkpoint based on dev set
F1 scores according to the ``exact'' metrics.\footnote{
Even though we report recall on GENIA,
model selection is still performed using F1.
}
For the \bert feature extractor,
we finetune the pretrained case-sensitive \bertbase model
through the \texttt{transformers} package.\footnote{
\url{github.com/huggingface/transformers}
}
For the non-\bert model,
we use pre-trained GloVe embeddings \citep{pennington+14}.

Following prior practice,
we embed gold POS tags as input features when using bi-LSTM for
the models trained on the GENIA dataset,
but we omit the POS tag embeddings for the PTB dataset.

The training process for each model takes roughly $10$ hours
using an RTX 2080 Ti GPU; model inference speed is $41.9$ sentences per second.\footnote{
We have not yet done extensive optimization regarding GPU batching for greedy transition-based parsers.
}

\begin{table}[t]
    \small
    \centering
    \begin{tabular}{lc}
    \toprule
    \emph{Adam Optimizer:} & \\
    Initial learning rate for bi-LSTM & $10^{-3}$ \\
    Initial learning rate for \bert   & $10^{-5}$ \\
    $\beta_1$ & $0.9$ \\
    $\beta_2$ & $0.999$ \\
    $\epsilon$ & $10^{-8}$ \\
    Minibatch size & $8$ \\
    Linear warmup steps & $800$ \\
    Gradient clipping $L_2$ norm & $5.0$ \\
    \midrule

    \emph{Inputs to bi-LSTM:} & \\
    Word-embedding dimensionality & $100$ \\
    POS tag-embedding dimensionality & $32$ \\
    Character bi-LSTM layers & $1$ \\
    Character bi-LSTM dimensionality & $128$ \\
    \midrule

    \emph{Bi-LSTM:} & \\
    Number of layers & $3$ \\
    Dimensionality & $800$ \\
    Dropout & $0.3$ \\
    \midrule

    \emph{MLPs (same for all MLPs):} \\
    Number of hidden layers & $1$ \\
    Hidden layer dimensionality & $400$ \\
    Activation function & ReLU \\
    Dropout & 0.3 \\
    \bottomrule
    \end{tabular}

    \caption{Hyperparameters of our models.}
    \label{tbl:hyperparam}
\end{table}

We select our hyperparameters by hand.
Due to computational constraints,
our hand-tuning has been limited to setting the dropout rates,
and from the candidates set of  $\{0.0,0.1,0.3,0.5\}$ we chose $0.3$ 
based on dev-set performance.
Our hyperparameters are listed in \autoref{tbl:hyperparam}.

\section{Extended Results}
\label{app:results}

\begin{table*}[t]
\small
\centering
\begin{tabular}{@{\hspace{2pt}}l@{\hspace{5pt}}l|cccccc|cccccc}
\toprule
& & \multicolumn{6}{c|}{Dev} & \multicolumn{6}{c}{Test} \\
& & \multicolumn{3}{c}{All} & \multicolumn{3}{c|}{NP} & \multicolumn{3}{c}{All} & \multicolumn{3}{c}{NP} \\
& & P & R & F & P & R & F & P & R & F & P & R & F \\
\midrule
\multirow{4}{*}{Exact}
& \citetalias{teranishi+17} &
$74.13$ & $73.34$ & $73.74$ & $76.21$ & $75.51$ & $75.86$ & $71.48$ & $70.70$ & $71.08$ & $75.20$ & $74.84$ & $75.01$ \\
& \citetalias{teranishi+19} &
$76.95$ & $76.76$ & $76.85$ & $78.11$ & $77.57$ & $77.84$ & $75.33$ & $75.61$ & $75.47$ & $77.95$ & $77.70$ & $77.83$ \\
& Ours &
$77.19$ & $77.00$ & $77.10$ & $80.00$ & $79.63$ & $79.82$ & $76.62$ & $76.34$ & $76.48$ & $81.89$ & $81.37$ & $81.63$ \\
& ~~+\bert &
$84.25$ & $84.55$ & $84.40$ & $88.53$ & $88.33$ & $88.43$ & $83.85$ & $83.62$ & $83.74$ & $85.33$ & $85.19$ & $85.26$ \\
\midrule
\multirow{5}{*}{Inner}
& \citetalias{ficler-goldberg16} &
$72.34$ & $72.25$ & $72.29$ & $75.17$ & $74.82$ & $74.99$ & $72.81$ & $72.61$ & $72.70$ & $76.91$ & $75.31$ & $76.10$ \\
& \citetalias{teranishi+17} &
$76.04$ & $75.23$ & $75.63$ & $77.82$ & $77.11$ & $77.47$ & $74.14$ & $73.33$ & $73.74$ & $77.44$ & $77.07$ & $77.25$ \\
& \citetalias{teranishi+19} &
$79.19$ & $79.00$ & $79.10$ & $80.64$ & $80.09$ & $80.36$ & $77.60$ & $77.88$ & $77.74$ & $80.19$ & $79.93$ & $80.06$ \\
& Ours &
$78.61$ & $78.42$ & $78.51$ & $82.07$ & $81.69$ & $81.88$ & $78.45$ & $78.16$ & $78.30$ & $84.29$ & $83.76$ & $84.03$ \\
& ~~+\bert &
$85.19$ & $85.50$ & $85.34$ & $89.45$ & $89.24$ & $89.35$ & $84.58$ & $84.35$ & $84.46$ & $86.28$ & $86.15$ & $86.22$ \\
\bottomrule
\end{tabular}
\caption{
    Precision, recall, and F1 scores on the PTB dev and test sets.
}
\label{tbl:ptb-full}

\end{table*}

\begin{table*}[t]
\small
\centering
\begin{tabular}{llcccccccccc}
\toprule
& & NP & VP & ADJP & S & PP & UCP & SBAR & ADVP & Others & All\\
& Count & $2{,}317$ & $465$ & $321$ & $188$ & $167$ & $60$ & $56$ & $21$ & $3$ & $3{,}598$ \\
\midrule
\multirow{4}{*}{Exact}
& \citetalias{teranishi+17} &
$57.14$ & $54.83$ & $72.27$ & $~~8.51$ & $55.68$ & $28.33$ & $57.14$ & $85.71$ & $~~0.00$ & $55.22$ \\
& \citetalias{teranishi+19} &
$59.21$ & $64.94$ & $78.19$ & $53.19$ & $55.68$ & $48.33$ & $66.07$ & $90.47$ & $~~0.00$ & $61.22$ \\
& Ours &
$68.28$ & $58.71$ & $86.29$ & $56.38$ & $55.09$ & $51.67$ & $58.93$ & $95.24$ & $~~0.00$ & $67.09$ \\
& ~~+\bert &
$79.41$ & $76.34$ & $88.79$ & $77.13$ & $73.05$ & $61.67$ & $76.79$ & $100.0$ & $33.33$ & $79.18$ \\
\midrule
\multirow{6}{*}{Whole}
& \citetalias{hara+09} &
$64.2~~$ & $54.2~~$ & $80.4~~$ & $22.9~~$ & $59.9~~$ & $36.7~~$ & $51.8~~$ & $85.7~~$ & $66.7~~$ & $61.5~~$ \\
& \citetalias{ficler-goldberg16} &
$65.08$ & $71.82$ & $74.76$ & $17.02$ & $56.28$ & $51.67$ & $91.07$ & $80.95$ & $33.33$ & $64.14$ \\
& \citetalias{teranishi+17} &
$67.19$ & $63.65$ & $76.63$ & $53.19$ & $61.67$ & $35.00$ & $78.57$ & $85.71$ & $33.33$ & $66.31$ \\
& \citetalias{teranishi+19} &
$59.30$ & $65.16$ & $78.19$ & $53.19$ & $55.68$ & $48.33$ & $66.07$ & $90.47$ & $~~0.00$ & $61.31$ \\
& Ours &
$69.40$ & $59.35$ & $87.85$ & $57.45$ & $56.89$ & $51.67$ & $62.50$ & $95.24$ & $~~0.00$ & $68.23$ \\
& ~~+\bert &
$80.58$ & $76.77$ & $90.03$ & $78.19$ & $76.65$ & $61.67$ & $82.14$ & $100.0$ & $33.33$ & $80.41$ \\
\bottomrule
\end{tabular}
\caption{
    Recall on the GENIA dataset.
}
\label{tbl:genia-full}

\end{table*}

\autoref{tbl:ptb-full} and \autoref{tbl:genia-full}
include detailed evaluation results on the PTB and GENIA datasets.

\putbib[ref]
\end{bibunit}

\end{appendices}

\end{document}